\documentclass[10pt,twocolumn,letterpaper]{article}

\usepackage{wacv}              
\usepackage[accsupp]{axessibility}  

\definecolor{wacvblue}{rgb}{0.21,0.49,0.74}
\usepackage[pagebackref,breaklinks,colorlinks,allcolors=wacvblue]{hyperref}
\usepackage{tabularx, booktabs, xcolor, adjustbox}
\usepackage{booktabs}          
\usepackage{multirow}          
\usepackage[table]{xcolor}     
\usepackage{tabularx}          
\usepackage{siunitx}           
\newpage
\usepackage[table]{xcolor} 
\usepackage{graphicx}
\usepackage{booktabs}
\usepackage{multirow}
\usepackage{xcolor}
\usepackage[table]{xcolor}   
\usepackage{amsmath}   
\usepackage{mathabx}   
\usepackage{fmtcount}

\usepackage{pgf} 
\newcommand{\hmin}[1]{%
  \pgfmathsetmacro{\H}{int(#1)}%
  \pgfmathsetmacro{\M}{int(round(60*(#1-\H)))}%
  \ifnum\M=60\relax
    \pgfmathsetmacro{\H}{int(\H+1)}%
    \pgfmathsetmacro{\M}{0}%
  \fi
  \H\,h~\M\,min%
}

\def\wacvPaperID{124} 
\def\confName{WACV}
\def\confYear{2026}

\title{Low‑Rank Expert Merging for Multi‑Source Domain Adaptation \\ in Person Re‑Identification}

\author{
Taha Mustapha Nehdi, Nairouz Mrabah,  Atif Belal, Marco Pedersoli, Eric Granger\\
LIVIA, ILLS, Dept. of Systems Engineering, ETS Montreal, Canada\\
{\tt\small 
taha-mustapha.nehdi.1@ens.etsmtl.ca, nairouz.mrabah@gmail.com, atif.belal.1@ens.etsmtl.ca,}\\
{\tt\small \{marco.pedersoli, eric.granger\}@etsmtl.ca}
}

\begin{document}
\maketitle
\begin{abstract}
Adapting person re-identification (reID) models to new target environments remains a challenging problem that is typically addressed using unsupervised domain adaptation (UDA) methods. Recent works show that when labeled data originates from several distinct sources (e.g., datasets and cameras), considering each source separately and applying multi-source domain adaptation (MSDA) typically yields higher accuracy and robustness compared to blending the sources and performing conventional UDA. 
However, state-of-the-art MSDA methods learn domain-specific backbone models or require access to source domain data during adaptation, resulting in significant growth in training parameters and computational cost. 
In this paper, a \textbf{S}ource‑free \textbf{A}daptive \textbf{G}ated \textbf{E}xperts \textbf{(SAGE‑reID)} method is introduced for person reID. Our SAGE‑reID is a cost-effective, source-free MSDA method that first trains individual source-specific low-rank adapters (LoRA) through source-free UDA. Then, a lightweight gating network is introduced and trained to dynamically assign optimal merging weights for fusion of LoRA experts, enabling effective cross-domain knowledge transfer. 
While the number of backbone parameters remains constant across source domains, LoRA experts scale linearly but remain negligible in size ($\leqslant2\%$ per source), reducing both the memory consumption and risk of overfitting.
Extensive experiments conducted on three challenging benchmarks -- Market-1501, DukeMTMC-reID, and MSMT17 -- indicate that SAGE‑reID can outperform  state-of-the-art methods while remaining computationally efficient. Our code is available: \url{https://github.com/nehdiii/SAGE-reID}
\end{abstract}
    
\section{Introduction}
\label{sec:intro}

Person re-identification (reID) aims to match query images of individuals captured over a network of non-overlapping cameras against instances in a gallery. It has a wide-ranging applications in surveillance \cite{ye2021deep}, smart city \cite{khan2024deep}, and autonomous driving \cite{wong2020identifying}. 
\begin{figure}[t]              
  \centering
  \includegraphics[width=\linewidth,trim=5mm 4mm 5mm 4mm,clip]{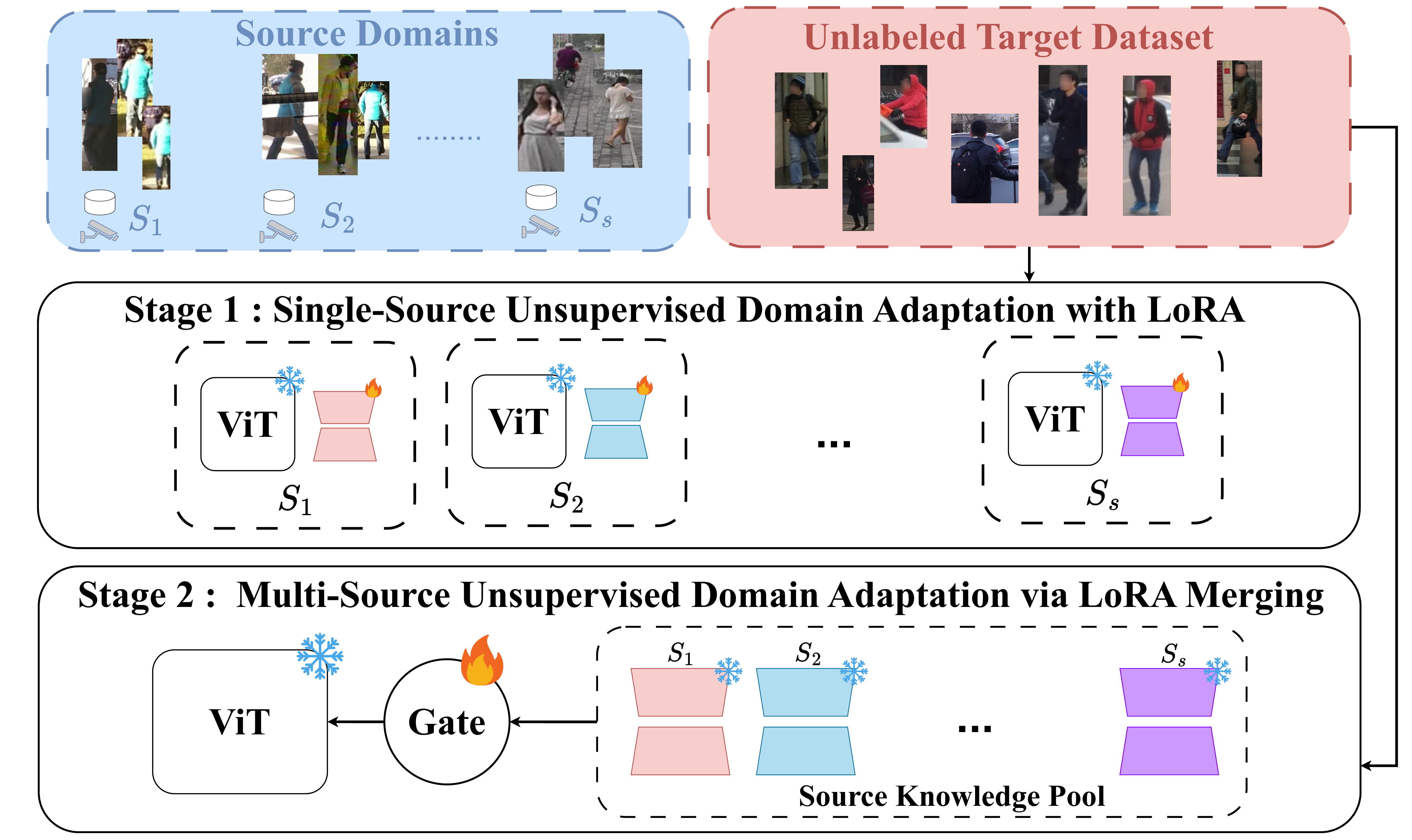}
  \caption{Illustration of our \textbf{SAGE‑reID} method. \textbf{Stage 1:} Single-source UDA with LORA: each pre-trained source-specific model is augmented with low-rank adapters, and is finetuned on the target domain. \textbf{Stage 2:} LoRA fusion: a lightweight gating network predicts a linear combination of weights for each adapter and merges them on-the-fly in a single forward pass.
 }
  \label{fig:overview}
  \label{fig:front}
\end{figure}
In recent years, supervised person reID systems have achieved remarkable progress \cite{Yang_2025_CVPR,Zhang_2023_CVPR,He_2021_ICCV,Li_2021_CVPR}. However, they often suffer from poor generalization when applied in unseen operational target domains due to the distribution shift between the source (lab training) and operational target (testing) data. Collecting large-scale labeled reID datasets is impractical due to the annotation cost. To mitigate this, several unsupervised domain adaptation (UDA) methods have been proposed \cite{Wei_2018_CVPR,zheng2021exploiting,Zheng_2021_CVPR,Lee_2023_ICCV,Guichemerre_2024_CVPR}, transferring knowledge from labeled source domains to unlabeled target domains. 
%
UDA methods for reID  \cite{Dai_2021_ICCV,ge2020mutual,ge2020selfpaced,he2022secret,Isobe_2021_ICCV,li2022reliability,zhao2020unsupervised,zheng2021exploiting,Zheng_2021_CVPR,zheng2021online,Lee_2023_ICCV} typically adapt a pretrained model on a single labeled source domain dataset and unlabeled target domain dataset. They only rely on supervision from only one source dataset, leaving the rich variation of other source domains untapped. 

MSDA for person reID was introduced \cite{Bai_2021_CVPR} to leverage supervision across varying environmental conditions, like background, camera viewpoint, illumination, and occlusion. By integrating complementary knowledge from multiple labeled source domains, MSDA produces more robust and discriminative target-domain representations than UDA approaches \cite{Bai_2021_CVPR,xian2025distilling}. 
%
Early MSDA methods assume that all domains share the same label set \cite{Peng_2019_ICCV,ganin2015unsupervised,yang2020curriculum, Chang_2019_CVPR}. However, in person reID, each dataset has distinct identities, so this shared-label premise fails and often causes negative transfer. To address this, RDSBN \cite{Bai_2021_CVPR} employs a two-stage pipeline: first, Rectified Domain-Specific Batch Normalization (RDSBN) adjusts each sample’s features to remove dataset bias; second, a graph-based Multi-Domain Information Fusion (MDIF) module merges knowledge from all source domains. However, MDIF requires access to source data  during adaptation, and this is often infeasible in real-world deployments due to privacy regulations, proprietary restrictions, and bandwidth constraints. CDR \cite{xian2025distilling} is the only source-free MSDA method for person reID. Given pre-trained expert models on each source domain, CDR adapts each expert to the target by self-training its mean-teacher model through clustering-based pseudo-labels. To ensure consistent features from each expert, a dual similarity consistency and decorrelation losses are used. However, its distillation stage requires full-backbone training, and during inference, each image is processed through all experts, resulting in substantial computational and memory costs that grow linearly with the number of sources.

To overcome these limitations, we introduce \textbf{S}ource‑free \textbf{A}daptive \textbf{G}ated \textbf{E}xperts for person reID \textbf{(SAGE‑reID)}, an MSDA framework for person reID. Starting from $s$ independently pre-trained source models, we adapt each one to the unlabeled target domain by only fine-tuning a lightweight low-rank adapter (LoRA) \cite{hu2022lora} appended to its frozen backbone (Fig.~\ref{fig:front}a). Pseudo-labels on the target domain are generated via DBSCAN clustering and are iteratively refined until convergence \cite{fan2018unsupervised}. These pseudo-labels are used to train each LoRA expert independently, aligning the model to the target distribution. In the subsequent domain-fusion stage, all backbone and LoRA parameters are kept frozen. A ``fused'' backbone is initialized by averaging the weights of $s$ source backbones corresponding to $s$ source domains, while each LoRA expert retains its target-adapted weights. A compact gating network then predicts a linear combination of weights per sample and performs only a single forward pass to merge the $s$ LoRA experts on-the-fly (Fig.~\ref{fig:front}b). This dynamic composition narrows the inter-domain gaps and reduces the computational and memory cost. 

The contributions of this paper are summarized as follows.
\textbf{(1)} A source-free MSDA method called \textbf{SAGE‑reID} is introduced that merges low-rank adapters to adapt each pre-trained expert to the target distribution, leveraging low-rank adapters to elegantly and efficiently capture target‑domain characteristics.  
\textbf{(2)} A lightweight gating module is designed that learns, in a "single" forward pass, an optimal linear combination of weights to merge the LoRA experts on-the-fly. 
\textbf{(3)} Extensive experiments are conducted on standard image-based person reID benchmarks, show that a significant gain in adaptation performance can be achieved with a cost-effective method.


%

\section{Related Works}
\label{sec:formatting}
\textbf{UDA for Person ReID}. 
UDA methods for person ReID aim to mitigate the domain gap between the source and the target domains. These methods can be categorized into two groups: generative methods and self-training. The first category of approaches \cite{Deng_2018_CVPR, Wei_2018_CVPR} employs generative models, like Generative Adversarial Networks (GANs) \cite{Goodfellow_2014_NeurIPS} to translate source images into target style images with identity-preserving techniques \cite{Deng_2018_CVPR}. The generated images are then used to train a re-ID model using supervised learning. These approaches often suffer from unstable training of GANs \cite{kodali2017convergence} and involve large parameter counts. 
The second line of work \cite{Dai_2021_ICCV, ge2020selfpaced, he2022secret, Isobe_2021_ICCV, li2022reliability, zhao2020unsupervised,  zheng2021exploiting, Zheng_2021_ICCV, Lee_2023_ICCV} relies on pseudo-label self-training, where a model trained on the source domain extracts features from the unlabeled target domain. These features are clustered, assigning pseudo-identity labels to each cluster. The quality of pseudo-labels significantly impacts re-ID performance on the target domain.
Several studies \cite{zhao2020unsupervised,  zheng2021exploiting, Zheng_2021_ICCV} refine these labels by ensembling multiple re-ID models and enforcing prediction consistency \cite{zhao2020unsupervised, zheng2021exploiting}; to mitigate camera bias, where clusters may contain images from a single camera, they employ cluster-weighting techniques \cite{Lee_2023_ICCV}. Rather than adapting in one step, \cite{Dai_2021_ICCV} progressively bridges domains by generating intermediate representations through cross-domain feature mixing \cite{verma2019manifold}. 
Despite their impressive performance, all UDA methods use only a single source domain, underutilizing the abundance of available labeled datasets.
%

\noindent\textbf{MSDA for Person ReID}. MSDA aims to adapt the model using multiple labeled source datasets and an unlabeled target dataset. Only two work addresses MSDA for person re-ID: MSUDA \cite{Bai_2021_CVPR} and CDR \cite{xian2025distilling}. MSUDA \cite{Bai_2021_CVPR} applies Rectification Domain-Specific Batch Norm (RDSBN) to neutralize dataset-specific artifacts and uses a graph convolution network to blend features via domain-agent nodes, yielding substantial gains over single-source baselines. However, MSUDA requires access to source data during adaptation, which might not be available due to privacy reasons. Source‑free MSDA (MSFDA) eliminates the need for source dataset access at adaptation time by leveraging only pre-trained source models and unlabeled target samples. CDR \cite{xian2025distilling} assigns each source domain its mean-teacher expert. After adapting each expert to the target domain, the features from all experts are concatenated to form a single embedding, and dual-similarity distillation and representation-decorrelation losses are employed to enforce complementary information. While \cite{xian2025distilling} achieves high performance, it trains all the parameters of each expert, causing computational cost to scale linearly with the number of sources. In contrast, our method only trains LoRA parameters, resulting in a significant reduction in computational overhead for the re-ID model regardless of the number of sources.

\noindent\textbf{Merging Low-Rank Experts}. Merging refers to the process of integrating separately fine‑tuned low‑rank adapters into a single model instance without full retraining. A conventional strategy for multi-skill learning involves training on a concatenation of datasets, often referred to as data mixing. While conceptually straightforward, this approach becomes increasingly impractical as the number of skills grows, since each new skill addition requires full model retraining on an updated mixture of datasets. To overcome this limitation, recent methods explore model merging as a scalable alternative. Rather than retraining, these techniques first independently fine-tune LoRA modules per skill and then merge them for composition. One family of approaches, known as the concatenation of LoRAs, linearly combines LoRA updates via layer-wise coefficients \cite{buehler2024x,feng2024mixture,luo2024moelora,muqeeth2024learning,wu2024mixture}. A second class of methods focuses on linear merging of LoRA parameters, which combines LoRA matrices before computing the final weight update. Examples include TIES \cite{yadav2023ties}  and DARE  \cite{yu2024language}, which incorporate sparsity and sign consistency to improve merge quality, and LoRA Hub \cite{huang2023lorahub}, which assigns static coefficients based on a few-shot target task. While these methods have primarily been studied for multi-task performance, we investigate their potential in MSDA.

\section{Proposed Method}

Consider \(s\) labeled source domains \(\{\mathcal{S}^{(i)}\}_{i=1}^{s}\) and an unlabeled target domain \(\mathcal{T}\).
The \(i^{\text{th}}\) source domain holds \(n_{S^{(i)}}\) image–label pairs \(\mathcal{S}^{(i)}=\{(x^{(i)}_j,y^{(i)}_j)\}_{j=1}^{n_{S^{(i)}}}\). The image $x^{(i)}_j$ denotes the $j^{\text{th}}$ observation for the $i^{\text{th}}$ source $\mathcal{S}^{(i)}$, and $y^{(i)}_j$ is its corresponding person identity. The target set contains \(n_T\) unlabeled samples \(\mathcal{T}=\{x_j\}_{j=1}^{n_T}\). 

In source-free adaptation, the source images are unavailable. Instead, the pretrained parameters of the \(s\) source models are used. Let \(L\) be the layer count and denote by \(\mathbf{W}_{0}^{(i,\ell)}\) the frozen weight matrix of layer \(\ell\) in source model \(i\), with \(i\!\in\!\{1,\dots,s\}\) and \(\ell\!\in\!\{1,\dots,L\}\). We define the parameters of backbone \(i\) as \(\mathcal{W}_0^{(i)}=\{\mathbf{W}_0^{(i,\ell)}\}_{\ell=1}^{L}\) and define the complete source prior as \(\mathbb{W}_0=\bigcup_{i=1}^{s}\mathcal{W}_0^{(i)}\). The goal is to exploit \(\mathbb{W}_0\) and \(\mathcal{T}\) to train a target model without revisiting any source data. 

\noindent \textbf{Low-rank adaptation.} LoRA~\cite{hu2022lora} augments each frozen weight matrix $\mathbf{W}_0$ with a trainable low-rank residual $\Delta\mathbf{W}$. The residual factorizes as $\Delta\mathbf{W}=\mathbf{A} \cdot \mathbf{B}$, where $\mathbf{A}\!\in\!\mathbb{R}^{n\times r}$ and $\mathbf{B}\!\in\!\mathbb{R}^{r\times m}$ with $r\!\ll\!\min(n,m)$. This low-rank factorization makes the adaptation fast and memory-efficient. The adapted weights are expressed as follows:  
\begin{equation}
\mathbf{W}= \mathbf{W}_0 + \mathbf{A} \cdot \mathbf{B}.
\end{equation}
During fine-tuning, only the parameters $\mathbf{A}$ and $\mathbf{B}$ of each layer are updated, while $\mathbf{W}_0$ stays fixed. This results in a parameter complexity of $\mathcal{O}\bigl(r(n+m)\bigr)$ for each layer, which is drastically lower than the $\mathcal{O}(nm)$ cost of full fine-tuning.

 \begin{figure*}[ht]
  \centering
  \includegraphics[width=0.86\linewidth]{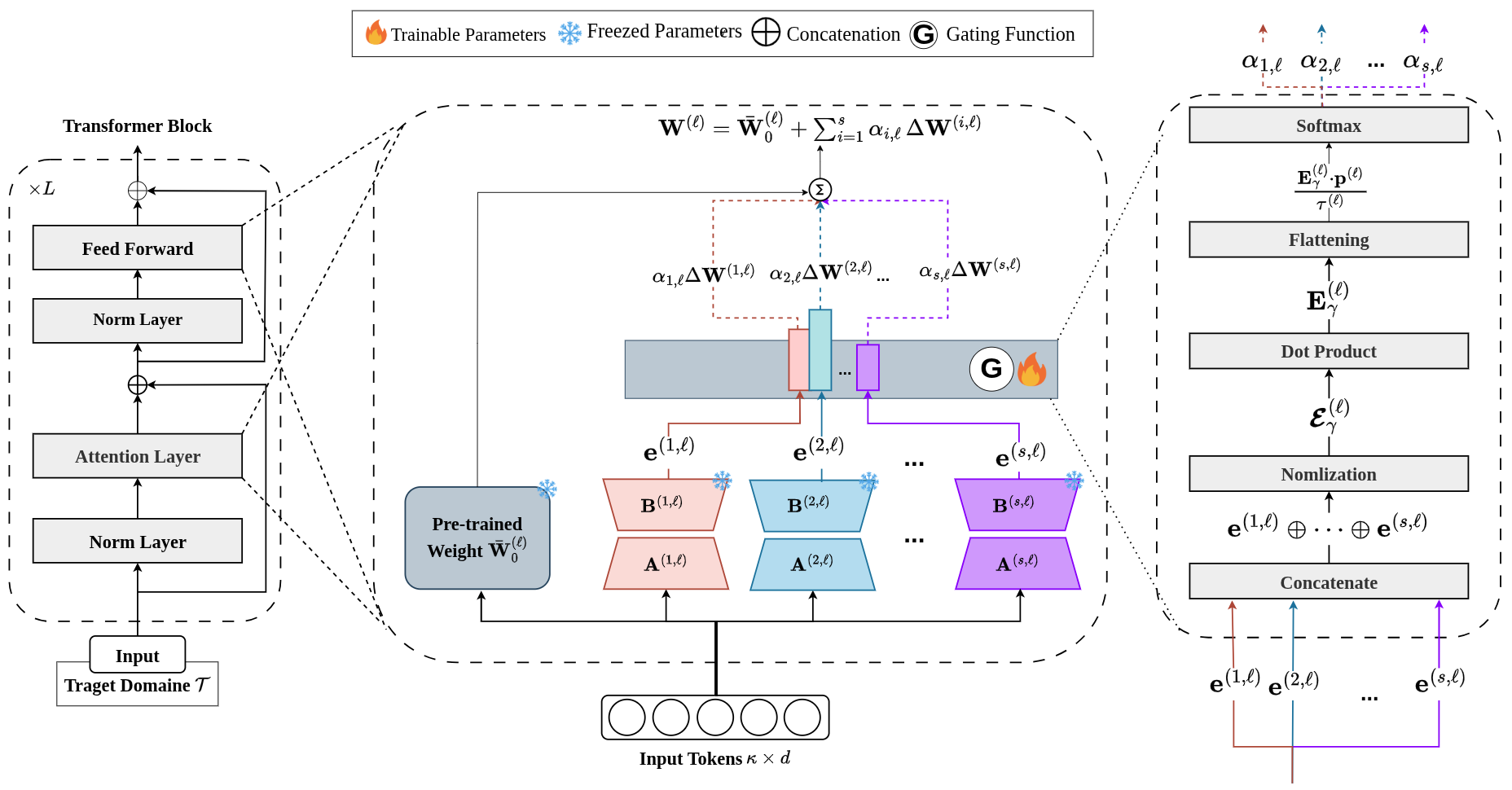}
   \caption{Illustration of the proposed low-rank expert merging stage (Stage 2) in our MSDA framework. Each source-specific LoRA expert, trained independently in Stage 1, provides a low-rank residual update. During training, a lightweight gating network dynamically predicts linear combination of coefficients for each expert at every transformer layer based on the input features. These coefficients are then used to combine the residuals, enabling fusion of multi-source knowledge through a single forward pass. The backbone and LoRA weights remain frozen, making this stage computationally light while enhancing generalization to the target domain.}
   \label{fig:Overview}
\end{figure*}


\subsection{Training Overview.}

This subsection describes the two-stage pipeline of our method, SAGE‑reID. In the first stage, we start from the $s$ source backbones, each pre-trained with cross-entropy and triplet losses on its source domain. For every source backbone $i$, we freeze all weights $\mathcal{W}_{0}^{(i)}$ and attach one LoRA pair $(\mathbf{A}^{(i, \ell)},\; \mathbf{B}^{(i, \ell)})$ to every layer $\ell$. Then, source-free UDA is performed on $\mathcal{T}$ via pseudo-supervision similar to \cite{fan2018unsupervised}. Only the LoRA pairs are trained for each backbone, which produces $s$ lightweight experts aligned to the target domain. In the second stage, a shared backbone is formed by averaging the weights of the source models $\widebar{\mathcal{W}}_0=\{\widebar{\mathbf{W}}^{(\ell)}_{0}\}_{\ell=1}^{L}$, such that $\widebar{\mathbf{W}}^{(\ell)}_{0} = \tfrac{1}{s}\sum_{i=1}^{s}\mathbf{W}_0^{(i,\ell)}$. Then, we fuse the experts through a linear combination:
\begin{equation}
\Delta\widebar{\mathbf{W}}^{(\ell)} = \sum_{i=1}^{s}\alpha_{i, \ell} \; \Delta\widebar{\mathbf{W}}^{(i, \ell)} = \sum_{i=1}^{s}\alpha_{i, \ell} \;\mathbf{A}^{(i,\ell)} \cdot \mathbf{B}^{(i,\ell)},
\end{equation}
where $\alpha_{i, \ell} \in [0,1]$ is a learnable coefficient for the $i^{\text{th}}$ source and $\ell^{\text{th}}$ layer. A single gradient step on the UDA loss estimates these coefficients while keeping the shared backbone and all LoRA parameters fixed. Figure~\ref{fig:Overview} outlines the second stage of SAGE‑reID. The next subsections elaborate on these two stages in full detail.

\subsection{First Stage} 
Each source encoder is a Vision Transformer \(f_{\mathcal{W}_0^{(i)}}\) followed by a linear classifier \(g_{\Theta^{(i)}}:\mathbb{R}^{D}\!\rightarrow\!\mathbb{R}^{C^{(i)}}\),
where \(\boldsymbol{\Theta}^{(i)}\) denotes the classifier weights, \(D\) is the latent space dimension, and \(C^{(i)}\) the number of identities in \(\mathcal{S}^{(i)}\).
For an image \(x_{j}^{(i)}\), we extract the feature \(z_{j}^{(i)} = f_{\mathcal{W}_0^{(i)}}(x_{j}^{(i)})\in\mathbb{R}^{D}\). Backbone and classifier are trained to minimize the joint loss:
\begin{align}
\mathcal{L}_{\mathcal{S}^{(i)}} &=
\sum_{j=1}^{n_{S^{(i)}}}
\mathcal{L}_{\text{CE}}\!\bigl(g_{\Theta^{(i)}}(z_{j}^{(i)}),\,y_{j}^{(i)}\bigr)
\notag \\                                           
&\quad+\;
\lambda\!\!
\sum_{(a,p,n)\in\mathcal{P}^{(i)}}
\!\mathcal{L}_{\text{Tri}}\!\bigl(z_{a}^{(i)},z_{p}^{(i)},z_{n}^{(i)}\bigr),
\label{eq:source}                                   
\end{align}

\begin{equation*}
\mathcal{L}_{\text{CE}}(\mathbf{u},y)=
-\log\frac{\exp(u_{y})}{\sum_{c=1}^{C^{(i)}}\exp(u_{c})},
\qquad
\mathbf{u}=g_{\Theta^{(i)}}(z),
\end{equation*}
\begin{equation*}
\mathcal{L}_{\text{Tri}}(z_{a},z_{p},z_{n})=
\max\!\bigl\{0,\;m+\lVert z_{a}-z_{p}\rVert_{2}-\lVert z_{a}-z_{n}\rVert_{2}\bigr\},
\end{equation*}
where \(\mathcal{P}^{(i)}\) is the set of mined anchor–positive–negative triplets, the scalar $m \in \mathbb{R}^{*}_{+}$ is the margin value, and the scalar \(\lambda \in \mathbb{R}_{+}\) balances classification and metric objectives.

After pre-training, we append a trainable low‑rank adapter \(\Delta\mathbf{W}^{(i,\ell)}=\mathbf{A}^{(i,\ell)} \cdot \mathbf{B}^{(i,\ell)}\), with rank \(r\ll\min(n,m)\), for every layer \(\ell\). The adapted model after integrating the low-rank modules is denoted \(f_{\,\mathcal{W}_0^{(i)}+\Delta\mathcal{W}^{(i)}}\), where \(\Delta\mathcal{W}^{(i)}=\{\Delta\mathbf{W}^{(i,\ell)}\}_{\ell=1}^{L}\). For every backbone \(i\), we freeze \(\mathcal{W}_{0}^{(i)}\) and update $\Delta\mathcal{W}^{(i)}$ and the classification head.

At each epoch, we extract target features \(z_{j}^{(i)} = f_{\,\mathcal{W}_0^{(i)}+\Delta\mathcal{W}^{(i)}}(x_{j})\) using backbone $i$ and cluster them with DBSCAN to obtain pseudo‑labels \(\hat{y}_{j}^{(i)}\). The low-rank adapters are trained to minimize the unsupervised losses:
\begin{align}
\mathcal{L}_{\text{UDA}}^{(i)} &=
\sum_{j=1}^{n_T}
\mathcal{L}_{\text{CE}}\!\bigl(g_{\Theta^{(i)}_{\mathcal{T}}}(z_{j}^{(i)}),\;\hat{y}_{j}^{(i)}\bigr)
\\
&+\;
\lambda\!\!
\sum_{(a,p,n)\in\hat{\mathcal{P}}^{(i)}}
\mathcal{L}_{\text{Tri}}\!\bigl(z_{a}^{(i)},z_{p}^{(i)},z_{n}^{(i)}\bigr),\nonumber
\label{eq:uda}
\end{align}
where \(\hat{\mathcal{P}}^{(i)}\) is the set of anchor–positive–negative triplets formed under the current pseudo‑labels for backbone $i$, and $\Theta^{(i)}_{\mathcal{T}}$ denotes the target classification head for backbone $i$. Each expert keeps its source prior while gradually aligning to the target distribution through the low‑rank residuals.


\subsection{Second Stage} \label{second_stage}
After the \(s\) experts finish target adaptation, we build a shared backbone by averaging the frozen source weights:
\begin{equation}
\widebar{\mathbf{W}}_{0}^{(\ell)}=\frac{1}{s}\sum_{i=1}^{s}\mathbf{W}_{0}^{(i,\ell)},\qquad
\widebar{\mathcal{W}}_{0}=\bigl\{\widebar{\mathbf{W}}_{0}^{(\ell)}\bigr\}_{\ell=1}^{L}.
\label{eq:merge_backbones}
\end{equation}
This shared backbone prevents drift when the low‑rank residuals are merged in the next step.

\noindent\textbf{Low‑rank Expert Merging.} The LoRA experts capture complementary biases toward the target domain. We fuse them through a lightweight, input‑adaptive gate. For each layer $\ell$, we keep the averaged backbone weight $\widebar{\mathbf{W}}_{0}^{(\ell)}$ and LoRA parameters frozen. Then, we combine the experts through the coefficients $\alpha_{i,\ell}\!\in\![0,1]$ with $\sum_{i=1}^{s}\alpha_{i,\ell}=1$. Thus, we obtain the weight matrix $\mathbf{W}^{(\ell)}$:
\begin{equation}
\mathbf{W}^{(\ell)}=
\widebar{\mathbf{W}}_{0}^{(\ell)}
+\sum_{i=1}^{s}\alpha_{i,\ell}\,\Delta\mathbf{W}^{(i,\ell)},
\label{eq:merge_layer}
\end{equation}
where \(\Delta\mathbf{W}^{(i,\ell)}=\mathbf{A}^{(i,\ell)} \cdot \mathbf{B}^{(i,\ell)}\) denotes the low‑rank update provided by expert \(i\) at layer \(\ell\).

\noindent\textbf{Gating Mechanism.} Let the input activation to layer $\ell$ through the shared backbone be $\mathbf{x}^{[\ell]}\in\mathbb{R}^{\kappa \times d}$ (we omit the batch index for clarity), with $\kappa$ denoting the number of tokens and $d$ the space dimension. Each expert produces a residual response $\mathbf{e}^{(i,\ell)}$ expressed as follows:
\begin{equation}
\mathbf{e}^{(i,\ell)}\;=\; \mathbf{x}^{[\ell]} \cdot \Delta\mathbf{W}^{(i,\ell)} \in\mathbb{R}^{\kappa \times d}.
\end{equation}
To align the scale of residuals across experts trained on diverse sources, we concatenate and $\ell_{2}$‑normalize them:
\begin{equation}
\boldsymbol{\mathcal{E}}_{\gamma}^{(\ell)} = \text{Norm}\big( \mathbf{e}^{(1,\ell)} \oplus \cdots \oplus \mathbf{e}^{(s,\ell)} \big) \in \mathbb{R}^{s \times \kappa \times d},
\end{equation}
where $\oplus$ is the concatenation operation. The normalization operation is applied with respect to the feature dimension $d$. We then flatten each expert residual to form a matrix $\mathbf{E}_{\gamma}^{(\ell)} \in \mathbb{R}^{s \times (\kappa d)}$. Finally, we compute the gating coefficient vector $\boldsymbol{\alpha}_{\ell}$ for layer $\ell$ using a trainable projection vector $\mathbf{p}^{(\ell)} \in \mathbb{R}^{\kappa d}$ and a learnable temperature value $\tau^{(\ell)} \in \mathbb{R}_{+}^{*}$ as expressed:
\begin{equation}
\boldsymbol{\alpha}_{\ell} = \text{Softmax}\left( \frac{\mathbf{E}_{\gamma}^{(\ell)} \cdot \mathbf{p}^{(\ell)}}{\tau^{(\ell)}} \right) = \big[\alpha_{1,\ell}, \cdots, \alpha_{s,\ell}\big]  \in \mathbb{R}^{s}.
\end{equation}

\noindent\textbf{Gate Training.} To preserve both general and domain-specific knowledge, we freeze the shared backbone and LoRA parameters during the gate training process and optimize the gating coefficients \(\mathbf{p}^{(\ell)}\) and temperature \(\tau^{(\ell)}\) for each layer $\ell$ along with the target classification head $g_{\Theta_{\mathcal{T}}}$, where $\Theta_\mathcal{T}$ denotes the classifier weight matrix.

We denote the final model after averaging the backbones and integrating the gating mechanism as \(f_{\,\widebar{\mathcal{W}}}\), where \(\widebar{\mathcal{W}} = \{\mathbf{W}^{(\ell)}\}_{\ell=1}^{L}\) and the weight matrix \(\mathbf{W}^{(\ell)}\) for layer $\ell$ is defined in Eq. (\ref{eq:merge_layer}). The target features \(z_{j} = f_{\widebar{\mathcal{W}}}(x_{j})\) are extracted using the shared backbone and merged LoRAs, and cluster these features with DBSCAN to obtain pseudo‑labels \(\hat{y}_{j}\). The gating mechanism is trained to minimize the loss:
\begin{align}
\mathcal{L}_{\text{UDA}} &=
\sum_{j=1}^{n_T}
\mathcal{L}_{\text{CE}}\!\bigl(g_{\Theta_{\mathcal{T}}}(z_{j}),\;\hat{y}_{j}\bigr) \; \\
&+\; \lambda\!\!
\sum_{(a,p,n)\in\hat{\mathcal{P}}}\mathcal{L}_{\text{Tri}}\!\bigl(z_{a},z_{p},z_{n}\bigr), \nonumber
\label{eq:uda_target}
\end{align}
where \(\hat{\mathcal{P}}\) is the set of anchor–positive–negative triplets formed under the current pseudo‑labels. Because of the small number of parameters, the gate converges quickly, requiring training for only one epoch on the target dataset. Each layer learns its own gating parameters $(\mathbf{p}^{(\ell)}, \;\tau^{(\ell)})$. Thus, the model can adaptively combine expert contributions in a depth-aware fashion.

\noindent\textbf{Parameter Cost.} Unlike the expert‑specific gating of~\cite{wu2024mixture}, which flattens all dimensions in $\boldsymbol{\mathcal{E}}_{\gamma}^{(\ell)}$ and requires $s$ projections of dimension $s \kappa d$ per layer, we use a \emph{single} projection vector $\mathbf{p}^{(\ell)} \in \mathbb{R}^{\kappa d}$ per layer. This reduces the memory and computation to $\mathcal{O}(\kappa d)$ instead of $\mathcal{O}(s^{2} \kappa d)$ per layer while preserving sufficient expressiveness for multi‑source domain adaptation. 

\section{Results and Discussion}

\subsection{Experimental Methodology}

\noindent\textbf{Datasets and Evaluation Metrics.} Experiments are carried out on four widely used person reID benchmarks:
Market‑1501 (M) ~\cite{Zheng_2015_ICCV}, DukeMTMC‑reID (D) ~\cite{ristani2016performance}, CUHK03 (CU) ~\cite{Li_2014_CVPR}, and MSMT17 (MS) ~\cite{Wei_2018_CVPR}. Statistics of these datasets are provided in Appendix A (Suppl. Mat.). Performance is reported with mean Average Precision (mAP) and Cumulative Matching Characteristic (CMC) at Rank‑1, following the standard evaluation protocol.

\noindent\textbf{Implementation Details.} We use the Vision Transformer (ViT) \cite{dosovitskiy2020image} pre-trained on ImageNet-21K and fine-tuned on ImageNet-1K as a backbone network, following the protocol in \cite{He_2021_ICCV}. During the first stage, each backbone is trained on a source dataset, followed by a 1-D BatchNorm, an $L_2$-normalization layer, and a classification head. After that, LoRA adapters are inserted into all projection layers of each transformer block (query, key, value, output, up, and down projections). For Market‑1501 and DukeMTMC, the rank and scaling coefficient of the low-rank adapters are set to $r$=4 and $\beta$=16, respectively. These hyperparameters are set to $r$=8 and $\beta$=32 for MSMT17. A LoRA dropout of 0.05 is applied across all datasets. All backbones are trained for 70 epochs with SGD and a fixed learning rate of 0.008. Input images are resized to 256$\times$128 and augmented with random horizontal flip, random crop, and random erasing following the protocol in \cite{zhong2020random}. Each mini‑batch consists of 64 images from 16 identities (4 images per identity). The hyperparameters remain unchanged for the second stage, except for the number of epochs, which is reduced to one. All experiments are conducted on two A100 GPUs.

\noindent\textbf{Baselines Methods.}
Table~\ref{tab:compact_sources_restructured} groups the competing approaches into four tiers that progressively relax domain-adaptation assumptions: (\romannumeral 1) \textit{Source-Only (No Adaptation).} A ViT\textsubscript{B/16} backbone is trained on a single labelled source with the supervised loss $\mathcal{L}_{\mathcal{S}}$ of Eq. (3) and evaluated directly on the target domain. This baseline constitutes a lower bound because no adaptation is performed on the target; (\romannumeral 2) \textit{Single-Source UDA.} It includes classical unsupervised domain adaptation (UDA) methods, such as MMT~\cite{ge2020mutual}, UNRN~\cite{zheng2021exploiting}, GLT~\cite{zheng2021group}, IDM~\cite{Dai_2021_ICCV}, SpCL~\cite{ge2020selfpaced}, and PUL~\cite{fan2018unsupervised}. MMT, UNRN, GLT, and PUL adhere to a two‑stage paradigm, first pre‑training on a single labeled source dataset and then fine‑tuning on the unlabeled target. In contrast, IDM and SpCL employ joint training, optimizing on both source and target images within every training mini‑batch rather than running a separate source‑only pre‑training step.
We add a baseline, $\mathcal{L}_{\text{UDA}}$ + ViT\textsubscript{B/16} (LoRA), which freezes a ViT\textsubscript{B/16} backbone after source pretraining, inserts LoRA adapters, and optimises these adapters with the unsupervised loss of Eq. (4); (\romannumeral 3) \textit{Multi-Source UDA (MSDA).} Approaches such as MSUDA~\cite{Bai_2021_CVPR} and CDM~\cite{xian2025distilling} exploit multiple heterogeneous sources during training to capture complementary domain cues before adapting to the target. To isolate the benefit of multi‑source awareness, we include, blending + $\mathcal{L}_{\text{UDA}}$ + ViT\textsubscript{B/16} (LoRA), a baseline that simply pools all sources, pretrains on the merged dataset and then performs low-rank adaptation. Our full model, SAGE‑reID, adds the second‑stage merging and gating mechanism described in Sec.~\ref{second_stage}; (\romannumeral 4) \textit{Supervised Methods.} The final block reports fully-supervised ResNet-50 and ViT\textsubscript{B/16} trained directly on target labels. This category indicates the upper bound performance achievable with complete supervision.


\subsection{Comparison with State-of-the-Art Methods}

\noindent\textbf{MSMT17.} In this benchmark, Market1501 (M), DukeMTMC-reID (D), and CUHK03 (CU) are used as source domains, while MSMT17 (MS) is used as the target domain. This setup is particularly challenging due to the large scale of the target dataset. In the single‑source UDA setting, our LoRA‑equipped ViT attains 39.0\% mAP for M $\!\rightarrow\!$ MS and 41.0\% mAP for D $\!\rightarrow\!$ MS, exceeding IDM by 5.5 pp and 7.5 pp, respectively. When all three sources are leveraged simultaneously, the full SAGE‑reID pipeline achieves 44.1\% mAP and 69.8\% Rank‑1, outperforming MSUDA by +9.2 mAP and +5.1 Rank‑1. These results demonstrate that (i) parameter‑efficient LoRA adapters enable Vision Transformers to generalize in large‑scale re‑ID transfer, and (ii) the proposed expert‑merging stage effectively consolidates heterogeneous source knowledge.  

\noindent\textbf{DukeMTMC-reID.} In this setting, Market1501 (M), MSMT17 (MS), and CUHK03 (CU) datasets are used as source domains, while the DukeMTMC-reID (D) dataset is used as the target domain. As shown in Table~\ref{tab:compact_sources_restructured} (middle), under the UDA setting, replacing the ResNet50 backbone of PUL with a frozen ViT\textsubscript{B/16} + LoRA raises mAP from 60.1\% to 69.2\% for M $\!\rightarrow\!$ D and from 61.2\% to 71.2\% for MS $\!\rightarrow\!$ D). These results can be explained by LoRA capacity to alleviate overfitting and confirm the benefit of low‑rank adaptation even without cross‑domain fusion. In the \textit{source-free} MSDA setting, the complete SAGE‑reID pipeline yields 72.7\% mAP and 84.1\% Rank‑1 and improves over the previous SOTA method, CDM, by +1.9 mAP and +1.8 Rank‑1. This margin showcases the effectiveness of the proposed expert averaging and lightweight gating mechanism.

\noindent\textbf{Market1501.} In this setting, DukeMTMC-reID (D), MSMT17 (MS), and CUHK03 (CU) are used as the source domains, while the Market1501 (M) dataset serves as the target domain. This scenario is challenging due to the small scale of the target dataset, which can cause strong overfitting, particularly when no source images are available during adaptation. As shown in Table~\ref{tab:compact_sources_restructured} (left), replacing PUL’s ResNet‑50 with a frozen ViT\textsubscript{B/16} + LoRA lifts mAP from 72.4\% to 81.5\% for D $\rightarrow$ M and from 73.5\% to 82.6\% for MS $\rightarrow$ M. Rank‑1 shows similar gains. These results highlight the capacity of LoRA in reducing overfitting, particularly for a small training dataset. IDM (single‑source) and MSUDA (multi‑source) score higher than our methods, but both of them retain and replay the source images during adaptation. This huge extra data regularizes the model and counters overfitting on Market1501. In contrast, CDM and SAGE‑reID are source‑free. Among source‑free MSDA approaches, SAGE‑reID yields the best scores (83.7\% mAP and 93.0\% Rank‑1), outperforming CDM by +2.5 mAP. Furthermore, our UDA scores match IDM and trail MSUDA by only 2.3 mAP, confirming the effectiveness of the proposed LoRA‑gate strategy without relying on source data.
\begin{table*}
\centering
\scriptsize
\renewcommand{\arraystretch}{1.2}
\setlength{\tabcolsep}{4pt}
\resizebox{0.78\textwidth}{!}{\begin{tabular}{l||ccc|ccc|ccc}
\toprule
\textbf{Methods} &
\multicolumn{3}{c|}{\underline{\textbf{Target: Market1501}}} &
\multicolumn{3}{c|}{\underline{\textbf{Target: DukeMTMC-reID}}} &
\multicolumn{3}{c}{\underline{\textbf{Target: MSMT17}}} \\
& \textbf{Source} & \textbf{mAP} & \textbf{R-1} & \textbf{Source} & \textbf{mAP} & \textbf{R-1} & \textbf{Source} & \textbf{mAP} & \textbf{R-1} \\
\midrule
\multicolumn{10}{l}{\textit{\textbf{$\cdot$ Lower Bound: Source-Only (No Adaptation)}}} \\
\midrule
$\mathcal{L}_{\mathcal{S}}$ + ViT\textsubscript{B/16} &  D &  34.0 &  61.3 &  M &  42.4 &  60.9 &  M &  12.6 &  32.0 \\
$\mathcal{L}_{\mathcal{S}}$ + ViT\textsubscript{B/16} &  MS &  40.1 &  66.9 &  MS &  51.7 &  68.7 &  D &  13.1 &  34.5 \\
    $\mathcal{L}_{\mathcal{S}}$ + ViT\textsubscript{B/16} & D + MS + CU  & \textbf{49.5} & \textbf{75.3} & M + MS + CU & \textbf{56.0} & \textbf{71.7} & D + M + CU  & \textbf{21.0} & \textbf{46.4} \\

\midrule
\multicolumn{10}{l}{\textit{\textbf{$\cdot$ UDA Methods}}} \\
\midrule
MMT~\cite{ge2020mutual}      &  D &  71.2 &  87.7 &  M &  65.1 &  78.0 &  M &  24.0 &  50.1 \\
UNRN~\cite{zheng2021exploiting}   &  D &  78.1 &  91.9 &  M &  69.1 &  82.0 &  M &  34.2 &  65.2 \\
GLT~\cite{zheng2021group}   &  D &  79.5 &  92.2 &  M &  69.2 &  82.0 & M &  26.5 & 56.6 \\
IDM~\cite{Dai_2021_ICCV}  &  D &  \textbf{82.8} &  \textbf{93.2} &  M &  \textbf{70.5} & \textbf{83.6} &  M &  33.5 &  61.3 \\
SpCL~\cite{ge2020selfpaced}   &  -- &  -- &  -- &  -- &  -- &  -- &  M &  26.8 &  53.7 \\
PUL~\cite{fan2018unsupervised}    &  D &  72.4 &  87.7 &  M &  60.1 &  76.1 &  M &  12.0 &  27.5 \\

\rowcolor{gray!30}
$\mathcal{L}_{\text{UDA}}$ + ViT\textsubscript{B/16} (LoRA) &  D &  81.5 &  92.5 &  M &  69.2 &  81.1 &  M &  \textbf{39.0} & \textbf{63.5} \\
\midrule
SpCL~\cite{ge2020selfpaced}   &  MS &  77.5 &  89.7 &  -- &  -- &  -- &  -- &  -- &  -- \\
GLT~\cite{zheng2021group}   &  -- &  -- &  -- &  -- &  -- &  -- &  -- &  27.7 &  59.5 \\
PUL~\cite{fan2018unsupervised}   &  MS &  73.5 &  88.7 &  MS &  61.2 &  76.0 &  D &  12.8 &  29.5 \\

\rowcolor{gray!30}
$\mathcal{L}_{\text{UDA}}$ + ViT\textsubscript{B/16} (LoRA) &  MS &  \textbf{82.6} &  \textbf{92.8} &  MS &  \textbf{71.2} &  \textbf{82.5} &  D &  \textbf{41.0} & \textbf{65.8} \\


\midrule
\multicolumn{10}{l}{\textit{\textbf{$\cdot$ MSDA Methods}}} \\
\midrule
MSUDA~\cite{Bai_2021_CVPR}              & D + MS + CU  & \textbf{86.0} & \textbf{94.8} & M + MS + CU & \textbf{68.9} & \textbf{82.1} & D + M + CU  & \textbf{34.9} & \textbf{64.7} \\
\midrule
\multicolumn{10}{l}{\textit{\textbf{$\cdot$ MSFDA: \textit{Source-free} MSDA Methods}}} \\
\midrule
CDM~\cite{xian2025distilling}           & D + MS + CU  & 81.2 & 92.9 & M + MS + CU & 70.8 & 82.3 & D + M + CU  & 32.9 & 63.8 \\
\rowcolor{gray!30}
Blend + $\mathcal{L}_{\text{UDA}}$ + ViT\textsubscript{B/16} (LoRA) & D + MS + CU  & 80.5 & 92.3 & M + MS + CU & 65.1 & 79.6 & D + M + CU  & 42.7 & 67.4 \\
\rowcolor{gray!30}
SAGE-ReID & D + MS + CU  & \textbf{83.7} & \textbf{93.0} & M + MS + CU & \textbf{72.7} & \textbf{84.1} & D + M + CU  & \textbf{44.1} & \textbf{69.8} \\
\midrule
\multicolumn{10}{l}{\textit{\textbf{$\cdot$ Upper Bound: Supervised Methods}}} \\
\midrule
ALDER + ResNet50~\cite{zhang2021seeing}           & --     & \textbf{88.9} & \textbf{95.6} & -- & 78.9 & 89.9 & --   & 59.1  & 82.2 \\ 
DCAL  + ViT\textsubscript{B/16}~\cite{Zhu_2022_CVPR}             & --     & 87.5 & 94.7 & -- & \textbf{80.1} & \textbf{89.0}   & --   & \textbf{64.0} & \textbf{83.1} \\
\bottomrule
\end{tabular}}
\caption{Performance of SAGE-ReID against state-of-the-art adaptation methods on three targets: Market‑1501 (left), DukeMTMC‑reID (centre), and MSMT17 (right). The table groups approaches into \emph{Source‑Only}, \emph{Single‑Source UDA}, \emph{Multi‑Source UDA (MSDA)}, and \emph{Supervised Methods}. Best scores in each block are in \textbf{bold} and grey rows denote our methods.}

\label{tab:compact_sources_restructured}
\end{table*}

\subsection{Ablation Study}
\noindent\textbf{Importance of Source Knowledge Separation.} \label{sec:sep} A central challenge in MSDA is how to exploit source-domain knowledge so that each domain contributes effectively to the target domain. Table~\ref{tab:ablation_num_sources} includes four baselines and shows results with respect to the number of source datasets. The four baselines are: (\romannumeral 1) Blend + $\mathcal{L}_{\mathcal{S}}$ + ViT\textsubscript{B/16}, all source images are pooled, and a single ViT\textsubscript{B/16} is trained with the supervised loss $\mathcal{L}_{\mathcal{S}}$ without any target fine‑tuning; (\romannumeral 2) Avg + $\mathcal{L}_{\mathcal{S}}$ + ViT\textsubscript{B/16}, a separate backbone is trained on each source. The $s$ trained networks are merged by simple weight averaging without any target fine‑tuning; (\romannumeral 3) Blend + $\mathcal{L}_{\text{UDA}}$ + ViT\textsubscript{B/16} (LoRA), the blended model of (\romannumeral 1) is taken as frozen backbone. LoRA adapters are inserted and updated on the target via the unsupervised loss $\mathcal{L}_{\text{UDA}}$; (\romannumeral 4)  Avg + $\mathcal{L}_{\text{UDA}}$ + ViT\textsubscript{B/16} (LoRA), identical to (\romannumeral 3) but the averaged backbone of (\romannumeral 2) is used instead of blending.

When only supervised pretraining is allowed, averaging the backbones (row~2) surpasses the blending-based model (row~1) by a significant margin, e.g.\ +3.3\, mAP for two sources and +4\, mAP for three sources. These results confirm that training with the sources independently avoids negative interference. After unsupervised adaptation, both LoRA variants boost mAP by more than 25 pp in mAP over the supervised counterparts, showing that lightweight target tuning is crucial. As we can see from the same table, initializing the adapters from the averaged backbone (row~4) is consistently better than the blending-based model (row~3). Furthermore, blending can even degrade performance when the number of sources increases. This result confirms again the negative interference caused by blending the sources.

Our model (row~5) adds a gating mechanism to fuse the LoRA experts. This yields another gain of 0.7–1.4\, mAP over the best baseline and reaches 44.1\%/69.8\% (mAP/R‑1) with three sources. These results show the effectiveness of source separation followed by learned re‑composition.

\begin{table}
\centering
\scriptsize
\setlength{\tabcolsep}{3pt}
\renewcommand{\arraystretch}{1.1}  
\begin{tabularx}{\linewidth}{l|*{6}{>{\centering\arraybackslash}X}}
\hline
\multirow{2}{*}{\textbf{Methods}} &
\multicolumn{2}{c}{\textbf{D}} &
\multicolumn{2}{c}{\textbf{D+CU}} &
\multicolumn{2}{c}{\textbf{D+CU+M}} \\
\cline{2-7}
& \textbf{mAP} & \textbf{R-1} & \textbf{mAP} & \textbf{R-1} & \textbf{mAP} & \textbf{R-1} \\
\hline
Blend + $\mathcal{L}_{\mathcal{S}}$ + ViT\textsubscript{B/16} & 13.1 & 34.5 & 15.8 & 37.5 & 17.0 & 39.9 \\
Avg + $\mathcal{L}_{\mathcal{S}}$ + ViT\textsubscript{B/16} & \textbf{13.1} & \textbf{34.5} & \textbf{19.1} & \textbf{44.3} & \textbf{21.0} & \textbf{46.4} \\
\hline
Blend + $\mathcal{L}_{\text{UDA}}$ + ViT\textsubscript{B/16} (LoRA)  & 41.0 & 65.8 & 40.3 & 65.3 & 42.7 & 67.4 \\
Avg + $\mathcal{L}_{\text{UDA}}$ + ViT\textsubscript{B/16} (LoRA)   & 41.0 & 65.8 & 41.5 & 67.4 & 43.4 & 68.4 \\
\hline
\rowcolor{gray!20}
SAGE-ReID & \textbf{41.0} & \textbf{65.8} & \textbf{43.5} & \textbf{69.3} & \textbf{44.1} & \textbf{69.8} \\
\hline
\end{tabularx}
\caption{Impact of source knowledge separation on MSMT17 with different numbers of source domains using ViT\textsubscript{B/16} backbone.}
\label{tab:ablation_num_sources}
\end{table}

\noindent\textbf{Effectiveness of Low-Rank Adaptation.} Table~\ref{tab:effect_lora} disentangles the contribution of (i) the backbone architecture, (ii) parameter–efficient low-rank adapters, and (iii) the proposed gating fusion. All variants share the same two‑stage protocol: supervised pre‑training on the source set with $\mathcal{L}_{\mathcal S}$, followed by source–free adaptation on the target with the pseudo‑label loss $\mathcal{L}_{\text{UDA}}$. First, we observe that replacing the fully fine-tuned CNN baseline ($\mathcal{L}_{\text{UDA}}$ + ResNet50) with a fully fine‑tuned Vision Transformer ($\mathcal{L}_{\text{UDA}}$ + ViT\textsubscript{B/16}) yields large accuracy gains on every transfer path. For the challenging M $\!\rightarrow\!$ MS task, mAP rises from 12.0\,\% to 39.4\,\%, while for M $\!\rightarrow\!$ D, it jumps from 60.1\,\% to 68.1\,\%. These results confirm the stronger representation capacity of ViT even when source data are no longer accessible during adaptation. Second, inserting low-rank adapters and freezing the ViT weights (\,$\mathcal{L}_{\text{UDA}}$ + ViT\textsubscript{B/16}(LoRA)\,) further improves or matches the fully tuneable ViT in five out of the nine single‑source transfers. The benefit is most visible when the target is small: on MS $\!\rightarrow\!$ M mAP grows from 76.7\,\% to 82.6\,\%, and on CU $\!\rightarrow\!$ M from 79.1\,\% to 81.6\,\%. These results support the finding that LoRA mitigates over‑fitting by restricting the number of trainable parameters. Third, SAGE‑reID merges three LoRA experts with the lightweight gate of Sec.~\ref{second_stage}. The fused model surpasses the best single‑source LoRA on every benchmark: +3.1 mAP / +4.0 R‑1 on MSMT17, +1.1 mAP / +0.2 R‑1 on Market1501, and +1.5 mAP / +0.8 R‑1 on DukeMTMC. These steady but non‑trivial margins indicate that the gate learns to combine complementary domain‑specific adaptations, which can not be obtained by naive weight averaging.


\begin{table}[t]
\centering
\scriptsize
\setlength{\tabcolsep}{3pt}        
\renewcommand{\arraystretch}{1.1} 
\begin{tabularx}{\linewidth}{l|*{6}{>{\centering\arraybackslash}X}}
\hline
\multirow{2}{*}{\textbf{Methods}} &
\multicolumn{2}{c}{\textbf{M}} &
\multicolumn{2}{c}{\textbf{D}} &
\multicolumn{2}{c}{\textbf{CU}} \\
\cline{2-7}
& \textbf{mAP} & \textbf{R-1} & \textbf{mAP} & \textbf{R-1} & \textbf{mAP} & \textbf{R-1} \\
\hline
\multicolumn{1}{c}{\textbf{\textit{Target: MSMT17}}} \\ \midrule
$\mathcal{L}_{\text{UDA}}$ + ResNet50 & 12.0 & 27.5 & 12.8 & 29.5 & 11.5 & 26.9 \\
$\mathcal{L}_{\text{UDA}}$ + ViT\textsubscript{B/16} & 39.4 & 65.5 & 39.5 & 65.9 & 36.7 & 62.8 \\
$\mathcal{L}_{\text{UDA}}$ + ViT\textsubscript{B/16} (LoRA) & 39.0 & 63.5 & 41.0 & 65.8 & 38.3 & 64.4 \\
\hline
\multicolumn{3}{l}{\textbf{\textit{Multi-Source (M + D + CU):}}} \\ \hline
\rowcolor{gray!15}
SAGE-ReID                & \multicolumn{6}{c}{\textbf{44.1 \quad 69.8}} \\ 
\hline
\multirow{2}{*}{\textbf{Methods}} &
\multicolumn{2}{c}{\textbf{D}} &
\multicolumn{2}{c}{\textbf{CU}} &
\multicolumn{2}{c}{\textbf{MS}} \\
\cline{2-7}
& \textbf{mAP} & \textbf{R-1} & \textbf{mAP} & \textbf{R-1} & \textbf{mAP} & \textbf{R-1} \\
\hline
\multicolumn{1}{c}{\textbf{\textit{Target: Market1501}}} \\ \midrule
$\mathcal{L}_{\text{UDA}}$ + ResNet50 & 72.4 & 87.7 & 72.5 & 87.6 & 73.5 & 88.7 \\
$\mathcal{L}_{\text{UDA}}$ + ViT\textsubscript{B/16}            & 72.6 & 88.0 & 79.1 & 91.7 & 76.7 & 89.8 \\
$\mathcal{L}_{\text{UDA}}$ + ViT\textsubscript{B/16} (LoRA)     & 81.5 & 92.5 & 81.6 & 92.0 & 82.6 & 92.8 \\
\hline
\multicolumn{3}{l}{\textbf{\textit{Multi-Source (D + CU + MS):}}} \\ \hline
\rowcolor{gray!15}
SAGE-ReID                & \multicolumn{6}{c}{\textbf{83.7 \quad 93.0}} \\ 
\hline
\multirow{2}{*}{\textbf{Methods}} &
\multicolumn{2}{c}{\textbf{M}} &
\multicolumn{2}{c}{\textbf{CU}} &
\multicolumn{2}{c}{\textbf{MS}} \\
\cline{2-7}
& \textbf{mAP} & \textbf{R-1} & \textbf{mAP} & \textbf{R-1} & \textbf{mAP} & \textbf{R-1} \\
\hline
\multicolumn{1}{c}{\textbf{\textit{Target: DukeMTMC-reID}}} \\ \midrule
$\mathcal{L}_{\text{UDA}}$ + ResNet50                      & 60.1 & 76.1 & 60.3 & 76.1 & 61.2 & 76.0 \\
$\mathcal{L}_{\text{UDA}}$ + ViT\textsubscript{B/16}        & 68.1 & 80.7 & 65.4 & 78.8 & 68.1 & 80.9 \\
$\mathcal{L}_{\text{UDA}}$ + ViT\textsubscript{B/16} (LoRA)        & 69.2 & 81.1 & 70.4 & 82.4 & 71.2 & 82.5 \\
\hline
\multicolumn{3}{l}{\textbf{\textit{Multi-Source (M + CU + MS):}}} \\ \hline
\rowcolor{gray!15}
SAGE-ReID                 & \multicolumn{6}{c}{\textbf{72.7 \quad 83.3}} \\
\hline
\end{tabularx}
\caption{Impact of LoRA adapters, Vision Transformers, and the gating mechanism. All results are reported after adaptation, and no source images are used during fine‑tuning.}
\label{tab:effect_lora}
\end{table}


\noindent\textbf{Comparison with MSUDA.} MSUDA~\cite{Bai_2021_CVPR} proposes two architectural enhancements: (\romannumeral 1) RDSBN, an improved domain-specific batch-normalization layer that strengthens domain-specific learning, and (\romannumeral 2) MDIF, a feature-fusion module. Because the original model was trained with the MMT loss ~\cite{ge2020mutual}, we re-evaluate it with the loss function $\mathcal{L}_{\text{UDA}}$ for a fair comparison. The results are presented in Table~\ref{tab:faircomp}. Our method outperforms MSUDA by +3.3 mAP and +0.7 top-1 accuracy and +7.2 mAP and +4.1 top-1 accuracy in D + CU $\rightarrow$ M and M + CU $\rightarrow$ D, respectively, which shows the effectiveness of the proposed method.  

\begin{table}[t]
\centering
\scriptsize
\setlength{\tabcolsep}{3pt}
\renewcommand{\arraystretch}{1.1}  

\begin{tabularx}{\linewidth}{l|*{4}{>{\centering\arraybackslash}X}}
\hline
\multirow{2}{*}{\textbf{Methods}} &
\multicolumn{2}{c}{\textbf{D + CU $\rightarrow$ M}} &
\multicolumn{2}{c}{\textbf{M + CU $\rightarrow$ D}} \\
\cline{2-5}
& \textbf{mAP} & \textbf{R-1} & \textbf{mAP} & \textbf{R-1} \\
\hline
$\mathcal{L}_{\text{UDA}}$ + DSBN*      & 74.1 & 89.2 & 60.0 & 76.2 \\
$\mathcal{L}_{\text{UDA}}$ + MSUDA*     & 79.4 & 92.1 & 65.0 & 79.1 \\
\hline
\rowcolor{gray!20}
SAGE-ReID & \textbf{82.7} & \textbf{92.8} & \textbf{72.2} & \textbf{83.2} \\
\hline
\end{tabularx}
\caption{Multi‑source adaptation under $\mathcal{L}_{\text{UDA}}$ with a ViT‑B/16 backbone. *Results copied from the original MSUDA paper \cite{Bai_2021_CVPR}.}
\label{tab:faircomp}
\end{table}

\noindent\textbf{Comparison with LoRA Composition and Weight Merging.}
In Table~\ref{tab:merge_loras_reb}, we benchmark state-of-the-art merging families under our MSFDA protocol with identical ViT-B/16 backbones, $\mathcal{L}_{\text{UDA}}$ pseudo-supervision, and training budgets (all methods fuse the same target-adapted LoRA experts) on two (D + CU + M) $\rightarrow$ MS and (CU + M + MS) $\rightarrow$ D transfers. Task Arithmetic (TA)~\cite{ilharco2022editing} creates task vectors (difference between the finetuned and pretrained weights) then merges them by summation, yielding moderate results (33.5/59.4 on MS; 66.2/80.8 on D). TIES~\cite{NEURIPS2023_1644c9af} resolves sign conflicts in task-vector space but underperforms TA on MS (23.1/49.2) while remaining competitive on D (57.3/72.8). KnOTS~\cite{stoica2024model} observes that LoRA updates lie in a different subspace than full-weight task vectors. To address this problem, KnOTS aligns the LoRAs into a shared subspace, improving both TA and TIES (KnOTS--TA: 34.8/61.6 on MS, 67.3/80.5 on D; KnOTS--TIES: 24.8/51.6 on MS, 59.2/73.6 on D). TSV-M~\cite{Gargiulo_2025_CVPR} further reduces interference via singular-vector decorrelation between task vectors. Thus, the merged directions for TSV-M interfere less (36.6/62.9 on MS; 70.0/80.3 on D). 
Among LoRA-focused composition methods with few-shot target supervision (pseudo-labels in our case), LoRAHub~\cite{huang2023lorahub} learns global static and layer-shared coefficients and surpasses data-free approaches (40.4/67.4 on MS; 69.4/82.3 on D), while MoLE~\cite{wu2024mixture} performs per-token expert gating (i.e., for each token embedding, the gate predicts soft weights across LoRA experts) and yields strong results (43.8/69.4 on MS and 71.5/83.0 on D). However, our gating attains the best results on both transfers with a margin over MoLE of +1.2/+0.3 (mAP/R-1) on (CU+M+MS)$\rightarrow$D (Table~\ref{tab:merge_loras_reb}). The gains indicate that the input-aware composition of LoRA residuals is more effective than static or weight-space merges, while retaining favorable compute/parameter cost (see Suppl.\ Tables ~4 \& 5).

\begin{table}[t]
\centering
\scriptsize
\setlength{\tabcolsep}{1pt}
\begin{tabularx}{0.47\textwidth}{l c *{4}{>{\centering\arraybackslash}X}}
\toprule
\multirow{2}{*}{\textbf{Method}} & \multirow{2}{*}{\textbf{Venue}} 
& \multicolumn{2}{c}{\textbf{(D + CU + M) $\rightarrow$ MS}} 
& \multicolumn{2}{c}{\textbf{(CU + M + MS) $\rightarrow$ D}} \\
\cmidrule(lr){3-4}\cmidrule(lr){5-6}
& & \textbf{mAP} & \textbf{R-1} & \textbf{mAP} & \textbf{R-1} \\
\midrule

TIES                
& NeurIPS'24 
& 23.1 & 49.2 
& 57.3 & 72.8 \\

KnOTS–TIES         
& ICLR'25    
&  24.8   &  51.6   
& 59.2  & 73.6 \\

TA     
& ICLR'23    
& 33.5 & 59.4 
& 66.2 & 80.8 \\

KnOTS–TA           
& ICLR'25    
& 34.8 &  61.6   
&  67.3   & 80.5   \\

TSV-M               
& CVPR'25    
&  36.6   &  62.9   
& 70.0 & 80.3 \\

LoRAHub             
& COLM'24    
&  40.4   &  67.4   
& 69.4 & 82.3 \\

MoLE                
& ICLR'24    
&  43.8   &  69.4   
&  71.5   &  83.0   \\

\rowcolor{gray!15}
SAGE-ReID  
& \textbf{-} 
& \textbf{44.1} & \textbf{69.8} 
& \textbf{72.7} & \textbf{83.3} \\

\bottomrule
\end{tabularx}
\caption{LoRA and weight merging methods under our MSFDA protocol (all methods fuse the same target-adapted LoRA experts).}
\label{tab:merge_loras_reb}
\end{table}

\section{Conclusion}
Current MSDA methods for person re‑identification require access to source‑domain data during adaptation and train a full backbone for each source. In this paper, we introduced SAGE‑reID, a source‑free MSDA framework that fine‑tunes only lightweight low‑rank adapters (LoRA) per source model and leverages a compact, learnable gating network to fuse them on‑the‑fly in a single forward pass. This design requires training only a few parameters, significantly reducing the computational and memory costs, and eliminating the need for source data at adaptation time. Comprehensive experiments on Market‑1501, DukeMTMC‑reID, and MSMT17 demonstrate that SAGE‑reID consistently outperforms state‑of‑the‑art MSDA approaches in both accuracy and efficiency. A potential limitation is the naive averaging of source backbones for initialization. Future work will explore fully unsupervised initialization schemes.

\paragraph{Acknowledgements} This work was supported by Distech Controls Inc., the Natural Sciences and Engineering Research Council of Canada, the Digital Research Alliance of Canada, and MITACS.

\appendix  

\section{Datasets and Evaluation Metrics} Experiments are carried out on four widely used person reID benchmarks:
Market‑1501 (M) ~\cite{Zheng_2015_ICCV}, DukeMTMC‑reID (D) ~\cite{ristani2016performance}, CUHK03 (CU) ~\cite{Li_2014_CVPR}, and MSMT17 (MS) ~\cite{Wei_2018_CVPR}, IUSTPersonReID (IUST) \cite{moghaddam2025culturally}, Occluded-DukeMTMC (OC-D)\cite{Miao_2019_ICCV}. Statistics of these datasets are summarized in Table \ref{tab:dataset_stats}. Performance is reported with mean Average Precision (mAP) and Cumulative Matching Characteristic (CMC) at Rank‑1, following the standard protocol.

\begin{table*}[!ht]
  \centering
  \resizebox{\textwidth}{!}{%
  \begin{tabular}{l|cccccc}
    \toprule
    \textbf{Statistic} &
    \textbf{Market1501 \cite{Zheng_2015_ICCV}} &
    \textbf{DukeMTMC-reID \cite{ristani2016performance}} &
    \textbf{CUHK03 \cite{Li_2014_CVPR}} &
    \textbf{MSMT17 \cite{Wei_2018_CVPR}} &
    \textbf{IUSTPersonReID \cite{moghaddam2025culturally}} &
    \textbf{Occluded-DukeMTMC \cite{Miao_2019_ICCV}} \\
    \midrule
    \# Cameras                 & 6        & 8        & 2        & 15        & 19        & 8 \\[2pt]
    \# Images                  & 32,217   & 36,411   & 28,193   & 126,441   & 117,455   & 36,411 \\[2pt]
    \# IDs                     & 1,501    & 1,404    & 1,467    & 4,101     & 1,847     & 1221 \\[2pt]
    Train (img / IDs)          & 12,936 / 751 & 16,522 / 702 & 26,263 / 1,367 & 32,621 / 1,041 & 72,393 / 1,193 & 15,618 / 702 \\[2pt]
    Query (img / IDs)          & 3,368 / 750  & 2,228 / 702  & 200 / 100      & 11,659 / 3,060 & 1,428 / 654     & 2,210 / 519 \\[2pt]
    Gallery (img / IDs)        & 15,913 / 750 & 17,661 / 702 & 1,730 / 100    & 82,161 / 3,060 & 43,634 / 654    & 17,661 / 1,110 \\
    \bottomrule
  \end{tabular}}
  \caption{Statistics of the datasets used for training and evaluation. For Occluded-DukeMTMC, total images are computed as Train + Gallery + Query.}
  \label{tab:dataset_stats}
\end{table*}

\section{Multi-camera as Multi-Source Case Study} 

\begin{table}
\centering
\small
\setlength{\tabcolsep}{2pt}      
\renewcommand{\arraystretch}{1.12}

\resizebox{\linewidth}{!}{%
\begin{tabular}{l||ccc|ccc}
\toprule
        & \multicolumn{3}{c|}{\textbf{Cam 6}} & \multicolumn{3}{c}{\textbf{Cam 5}} \\
\cmidrule(lr){2-4}\cmidrule(lr){5-7}
\textbf{Source: Model} &
\textbf{mAP} & \textbf{R1} & $\mathbf{T_{\text{inf}}}$ &
\textbf{mAP} & \textbf{R1} & $\mathbf{T_{\text{inf}}}$ \\
\midrule
Cam 12: $\mathcal{L}_{\mathcal{S}}$ + ViT\textsubscript{B/16} & 73.5 & 92.2 & 0.10 & 71.4 & 87.1 & 0.10 \\
Cam 12: $\mathcal{L}_{\text{UDA}}$ + ViT\textsubscript{B/16} (LoRA) & 74.1 & 91.6 & 0.10 & 75.1 & 88.4 & 0.10 \\
\hline
Cam 13: $\mathcal{L}_{\mathcal{S}}$ + ViT\textsubscript{B/16} & 70.2 & 88.9 & 0.10 & 74.7 & 89.3 & 0.10 \\
Cam 13: $\mathcal{L}_{\text{UDA}}$ + ViT\textsubscript{B/16} (LoRA) & 71.1 & 88.9 & 0.10 & 76.2 & 89.8 & 0.10 \\
\hline
Cam 14: $\mathcal{L}_{\mathcal{S}}$ + ViT\textsubscript{B/16} & 74.9 & 91.4 & 0.10 & 79.3 & 91.2 & 0.10 \\
Cam 14: $\mathcal{L}_{\text{UDA}}$ + ViT\textsubscript{B/16} (LoRA)   & 76.1 & 91.2 & 0.10 & 81.0 & 91.8 & 0.10 \\
\hline
Cam 15: $\mathcal{L}_{\mathcal{S}}$ + ViT\textsubscript{B/16}     & 70.5 & 90.6 & 0.10 & 77.1 & 90.4 & 0.10 \\
Cam 15: $\mathcal{L}_{\text{UDA}}$ + ViT\textsubscript{B/16} (LoRA)    & 72.1 & 90.6 & 0.10 & 78.8 & 90.8 & 0.10 \\
\addlinespace[0.15em]
\hline
Cams 15\,+\,14: $\mathcal{L}_{\mathcal{S}}$ + ViT\textsubscript{B/16} & 78.8 & 92.8 & 0.26  & 82.2 & 92.8 & 0.26 \\
Cams 15\,+\,14: Blend + $\mathcal{L}_{\text{UDA}}$  + ViT\textsubscript{B/16} (LoRA)        & 78.6 & 91.8 & 0.10  & 82.2 & 91.9 & 0.10 \\
Cams 15\,+\,14: SAGE-ReID    & \textbf{79.5} & \textbf{93.2} & 0.26  & \textbf{83.9} & \textbf{92.9} & 0.26 \\
\hline
Cams 15\,+\,14\,+\,13: $\mathcal{L}_{\mathcal{S}}$ + ViT\textsubscript{B/16}       & 79.5 & 93.2 & 0.32  & 83.9 & 93.1 & 0.32 \\
Cams 15\,+\,14\,+\,13: Blend + $\mathcal{L}_{\text{UDA}}$  + ViT\textsubscript{B/16} (LoRA)             & 79.8 & 92.8 & 0.10  & 83.5 & 92.3 & 0.10 \\
Cams 15\,+\,14\,+\,13: SAGE-ReID       & \textbf{80.2} & \textbf{93.4} & 0.32 & \textbf{84.4} & \textbf{93.4} & 0.32 \\
\hline
Cams 15\,+\,14\,+\,13\,+\,12: $\mathcal{L}_{\mathcal{S}}$ + ViT\textsubscript{B/16}   & 80.2 & 93.4 & 0.36 & 83.2 & 92.8  & 0.36 \\
Cams 15\,+\,14\,+\,13\,+\,12: Blend + $\mathcal{L}_{\text{UDA}}$  + ViT\textsubscript{B/16} (LoRA)         & 80.1 & 93.4 & 0.10  & 83.5 & 92.6 & 0.10 \\
Cams 15\,+\,14\,+\,13\,+\,12: SAGE-ReID    & \textbf{80.8} & \textbf{93.8} & 0.36 & \textbf{83.9} & \textbf{93.1} & 0.36 \\
\bottomrule
\end{tabular}}
\caption{Per‑camera adaptation on MSMT17. Cams 12–15 are treated as source domains, while Cams 5 and 6 are unseen targets.}
\label{tab:cam_da_results}
\end{table}

A real-world scenario arises when a new camera is added to a pre-existing video surveillance network. 
This camera observes the scene from previously uncovered operating conditions (e.g., viewpoint and illumination). To emulate this scenario using the MSMT17 dataset, one of the largest and most challenging person re-ID benchmarks, containing over 15 camera subsets. Cameras 12–15 are designated as source domains, while cameras 6 and 5 as target domains. 
As we can see, averaging two source backbones, Cams 15+14 already surpasses the strongest single‑source baseline on Cam 5 (82.2 mAP vs. 79.3 mAP). These results indicate that complementary viewpoints are exploitable even without target data. Furthermore, unsupervised adaptation boosts every source set. For instance, Cam 12 $\rightarrow$ Cam 5 rises from 71.4$\%$ to 75.1$\%$ mAP (+3.7 pp). When multiple experts are available, naive source blending can decrease mAP and R-1 (e.g., Cams 15+14+13 $\rightarrow$ Cam 5). The proposed SAGE‑ReID gate instead improves performance to 84.4\% mAP / 93.4\% R‑1, and to 80.8\% mAP / 93.8\% R‑1 when four experts are fused for Cam 6. Latency scales modestly with the number of active experts, from 0.10ms (one camera) to 0.36ms (four cameras), well below a 30fps real‑time budget.

\section{Effectiveness of the Gating Mechanism} 
\begin{table}
\centering
\footnotesize
\setlength{\tabcolsep}{3pt}
\renewcommand{\arraystretch}{1.1}
\begin{tabularx}{\linewidth}{>{\raggedright\arraybackslash}X|c c}
\hline
& \multicolumn{2}{c}{\textbf{After Adaptation}} \\ \cline{2-3}
\textbf{Methods} & \textbf{mAP} & \textbf{R-1} \\ \hline
\multicolumn{3}{l}{\textbf{\textit{Multi-Source (M + D + CU) $\rightarrow$ MS:}}} \\ \hline
$\mathcal{L}_{\text{UDA}}$ + ViT\textsubscript{B/16} (LoRA) + LoRA Avg   & 43.2 & 69.5 \\
$\mathcal{L}_{\text{UDA}}$ + ViT\textsubscript{B/16} (LoRA) + LoRA MoLE \cite{wu2024mixture}  & 43.8 & 69.4 \\
\hline
\rowcolor{gray!20}
SAGE-ReID  & \textbf{44.1} & \textbf{69.8} \\ \hline
\multicolumn{3}{l}{\textbf{\textit{Multi-Source (D + CU + MS) $\rightarrow$ M:}}} \\ \hline
$\mathcal{L}_{\text{UDA}}$ + ViT\textsubscript{B/16} (LoRA) + LoRA Avg   & 79.5 & 91.9 \\
$\mathcal{L}_{\text{UDA}}$ + ViT\textsubscript{B/16} (LoRA) + LoRA MoLE \cite{wu2024mixture}  & 83.6 & 93.0 \\
\hline
\rowcolor{gray!20}
SAGE-ReID & \textbf{83.7} & \textbf{93.0} \\ \hline
\multicolumn{3}{l}{\textit{\textbf{Multi-Source (M + CU + MS) $\rightarrow$ D:}}} \\ \hline
$\mathcal{L}_{\text{UDA}}$ + ViT\textsubscript{B/16} (LoRA) + LoRA Avg & 71.7 & 83.1 \\
$\mathcal{L}_{\text{UDA}}$ + ViT\textsubscript{B/16} (LoRA) + LoRA MoLE \cite{wu2024mixture} & 71.5 & 83.0 \\
\hline
\rowcolor{gray!20}
SAGE-ReID & \textbf{72.7} & \textbf{83.3} \\ \hline
\end{tabularx}
\caption{Mean average precision (mAP) and Rank‑1 (R‑1) for three LoRA composition strategies after single-source adaptation.}
\label{tab:gateing_vs_MoLE}
\end{table}

\begin{table}
\centering
\footnotesize
\setlength{\tabcolsep}{4pt}
\renewcommand{\arraystretch}{1.1}
\begin{tabularx}{\linewidth}{c|>{\raggedright\arraybackslash}X|c c}
\hline
\# \textbf{Experts} & \textbf{Methods} & \textbf{GFLOPs / img} & \textbf{Params (M)} \\ \hline
\multirow{3}{*}{3}
  & LoRA Avg & 23.64 &  90.82 \\
  & LoRA MoLE & 23.86 & 187.12 \\
  \hline
  \rowcolor{gray!20}
  & SAGE-ReID             & 23.67 & 101.52 \\ \hline
\multirow{3}{*}{10}
  & LoRA Avg & 25.85 &  99.08 \\
  & LoRA MoLE & 28.09 & 1\,169.06 \\
  \hline
  \rowcolor{gray!20}
  & SAGE-ReID             & 25.95 & 109.78 \\ \hline
\end{tabularx}
\caption{Computational cost and parameter count of different LoRA‑composition schemes. LoRA Avg refers to $\mathcal{L}_{\text{UDA}}$ + ViT\textsubscript{B/16} (LoRA) + LoRA Avg and LoRA MoLE refers to $\mathcal{L}_{\text{UDA}}$ + ViT\textsubscript{B/16} (LoRA) + LoRA MoLE in Table \ref{tab:gateing_vs_MoLE}. “Params’’ is the \emph{total} number of parameters, including frozen ViT, LoRAs, and the gate module.}
\label{tab:compute_cost}
\end{table}
Table~\ref{tab:gateing_vs_MoLE} reports the proposed gating function performance versus two possible techniques. All variants first freeze the ViT\textsubscript{B/16} backbones trained on each source with $\mathcal{L}_{\mathcal S}$, then attach domain‑specific low-rank adapters trained on the target domain with the pseudo‑label loss $\mathcal{L}_{\text{UDA}}$. After that, all variants average the frozen source weights to build a shared backbone. Finally, we compare three ways to fuse the low-rank adapters: (\romannumeral 1) LoRA‑Avg, simply averages the low-rank adapters; (\romannumeral 2) LoRA‑MoLE, learns a per‑token mixture of experts as in \cite{wu2024mixture}; (\romannumeral 3) SAGE‑ReID employs the proposed gating mechanism as described in Sec. 3.2 of the main document.  
The proposed gate gives the best mAP on all transfers. On MSMT17, SAGE‑ReID improves over averaging by +0.9 mAP and over MoLE by +0.3 mAP. On DukeMTMC, SAGE‑ReID outperforms both baselines by more than +1.0 mAP. MoLE narrows the gap on the small Market‑1501 target but does so at the cost of a large gating tensor that scales quadratically with the number of sources $s$ and linearly with the sequence length $\kappa$. Table~\ref{tab:compute_cost} shows that for $s{=}10$ experts, MoLE inflates parameters by $\times11$, whereas our gate adds only +\,9 M parameters and +0.1 GFLOPs. The learnable parameters of our gate are of size $\kappa \times d$. Thus, the number of parameters and FLOPs stays fixed no matter how many sources there are. The shared gate, therefore, achieves better performance and cost while avoiding expert collapse observed with naive averaging. These results confirm that the proposed gate is effective and efficient to leverage complementary LoRA experts.



\section{Time \& Memory \& Parameter Efficiency}

We profile training-time and memory for the (M + D + CU) $\rightarrow$ MS transfer. Table~\ref{tab:model_profile} reports total/trainable number of parameters, training memory, per-iteration latency, and wall-clock time (batch size 32, $256{\times}128$, fp32).

\paragraph{Stage 1 (single-source target adaptation).} SAGE-ReID updates only LoRA modules, yielding 3.6M trainable parameters and 116.6\,MB training memory, whereas MSUDA/CDM updates the full backbone (76.8M; 1172.4\,MB). This provides a $>$10$\times$ reduction in training memory. Per-iteration latency is higher for SAGE-ReID (50.5\,ms vs.\ 13.5\,ms) and the wall-clock is longer (3\,h\,48\,min vs.\ 1\,h), but the parameter/memory footprint is substantially smaller.

\paragraph{Stage 2 (expert merging).} The proposed gate trains 10.6M parameters with 168.8\,MB training memory and converges in $\approx$1\,min. It is much lighter than MoLE (96.2M; 821.8\,MB; 148.0\,ms/iter) and far lighter and faster than MSUDA (72.4M; 1383.6\,MB; 5\,h). End-to-end, SAGE-ReID requires $\sim$3.8\,h vs.\ $\sim$6\,h for MSUDA.

\paragraph{Scalability and inference cost.} Our gate adds $\mathcal{O}(\kappa d)$ parameters per layer and does not grow with the number of sources $s$, while MoLE scales as $\mathcal{O}(s^{2}\kappa d)$. In Table \ref{tab:compute_cost}, with $s\!=\!10$ experts, MoLE’s parameters inflate to $\sim$1.17B ($\approx$11$\times$ SAGE-ReID), whereas our gate adds only $+9$M parameters and $+0.1$ GFLOPs; GFLOPs/image remain close across methods (e.g., 25.95 for SAGE-ReID). At inference time, SAGE-ReID performs a single merged forward pass.

\newcolumntype{L}{>{\raggedright\arraybackslash}X}
\newcolumntype{Y}{>{\centering\arraybackslash}m{8mm}} 
\sisetup{table-number-alignment=center}

\begin{table*}[t]
\centering
\footnotesize
\setlength{\tabcolsep}{4pt}
\renewcommand{\arraystretch}{1.05}
\begin{tabularx}{\textwidth}{@{}
  Y                         
  p{0.20\textwidth}         
  S[table-format=3.1]       
  S[table-format=3.1]       
  S[table-format=4.1]       
  S[table-format=3.1]       
  c                         
@{}}
\toprule
\multicolumn{1}{c}{} &
\textbf{Method} &
\multicolumn{1}{c}{\textbf{\# Total Params (M)}} &
\multicolumn{1}{c}{\textbf{\# Trainable Params (M)}} &
\multicolumn{1}{c}{\textbf{Trainable Memory (MB)}} &
\multicolumn{1}{c}{\textbf{ms/iter}} &
\textbf{Training Time (h/min)} \\
\midrule

\multirow{3}{*}{\rotatebox{90}{\textbf{Stage 1}}}
& SAGE-ReID & 88.4  &  3.6 &  116.6 &  50.5 & \hmin{3.8} \\
& MSUDA     & 76.8  & 76.8 & 1172.4 &  13.5 & \hmin{1.0} \\
& CDM       & 76.8  & 76.8 & 1172.4 & \multicolumn{1}{c}{--} & -- \\
\midrule

\multirow{4}{*}{\rotatebox{90}{\textbf{Stage 2}}}
& SAGE-ReID & 101.5 & 10.6 &  168.8 & 120.1 & \hmin{0.013} \\
& MoLE      & 187.1 & 96.2 &  821.8 & 148.0 & \hmin{0.016} \\
& MSUDA     & 144.8 & 72.4 & 1383.6 &  65.1 & \hmin{5} \\
& CDM       & 217.2 &108.6 & 1659.8 & \multicolumn{1}{c}{--} & -- \\
\bottomrule
\end{tabularx}
\caption{Model efficiency: latency is wall-clocked per iteration (batch size 32, image size $256\times128$, fp32). Training time is measured on MSMT17. M = million; MB = megabyte. CDM latency and total training time are not reported due to the unavailability of the code.}
\label{tab:model_profile}
\end{table*}

\section{Analysis of Gating Coefficients}
\label{sec:E}

\begin{figure*}[t]
  \centering
  \begin{subfigure}{\textwidth}
    \centering
    \includegraphics[height=0.28\textheight, keepaspectratio]{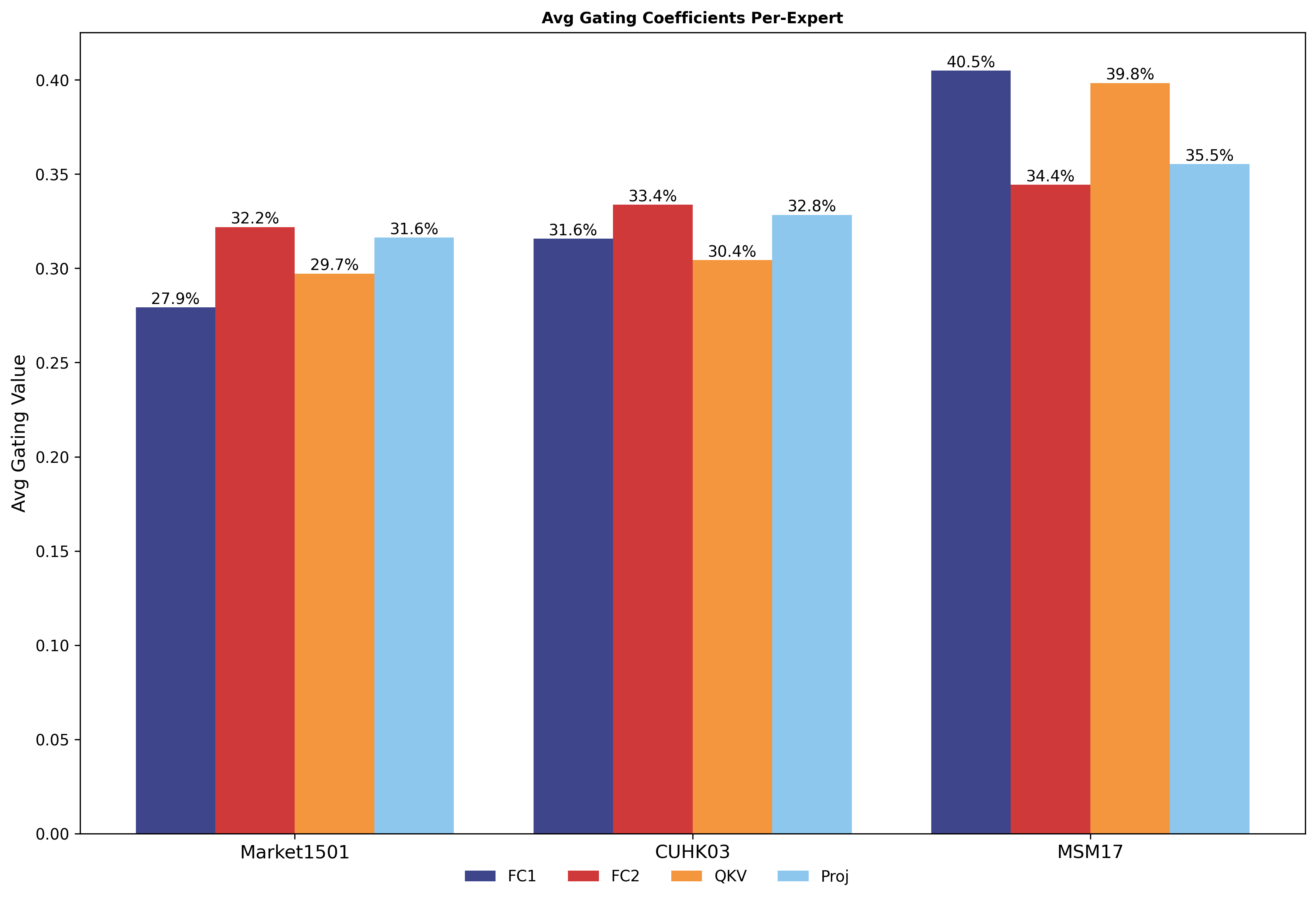}
    \subcaption{Average gating coefficient (Target: DukeMTMC-reID)}
    \label{fig:weights:duke}
  \end{subfigure}\vspace{4pt}

  \begin{subfigure}{\textwidth}
    \centering
    \includegraphics[height=0.28\textheight, keepaspectratio]{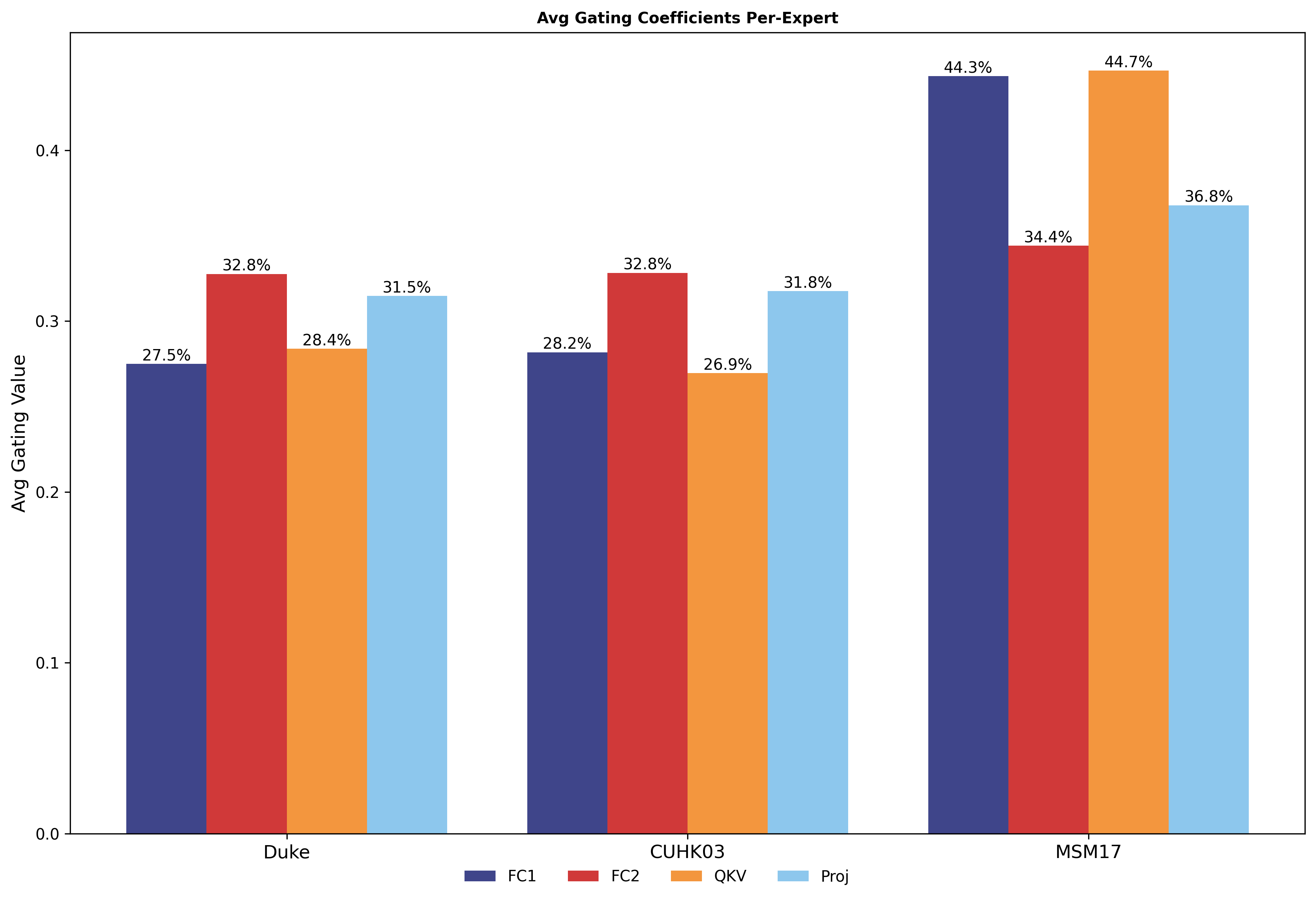}
    \subcaption{Average Average gating coefficient (Target: Market-1501)}
    \label{fig:weights:market}
  \end{subfigure}\vspace{4pt}

  \begin{subfigure}{\textwidth}
    \centering
    \includegraphics[height=0.28\textheight, keepaspectratio]{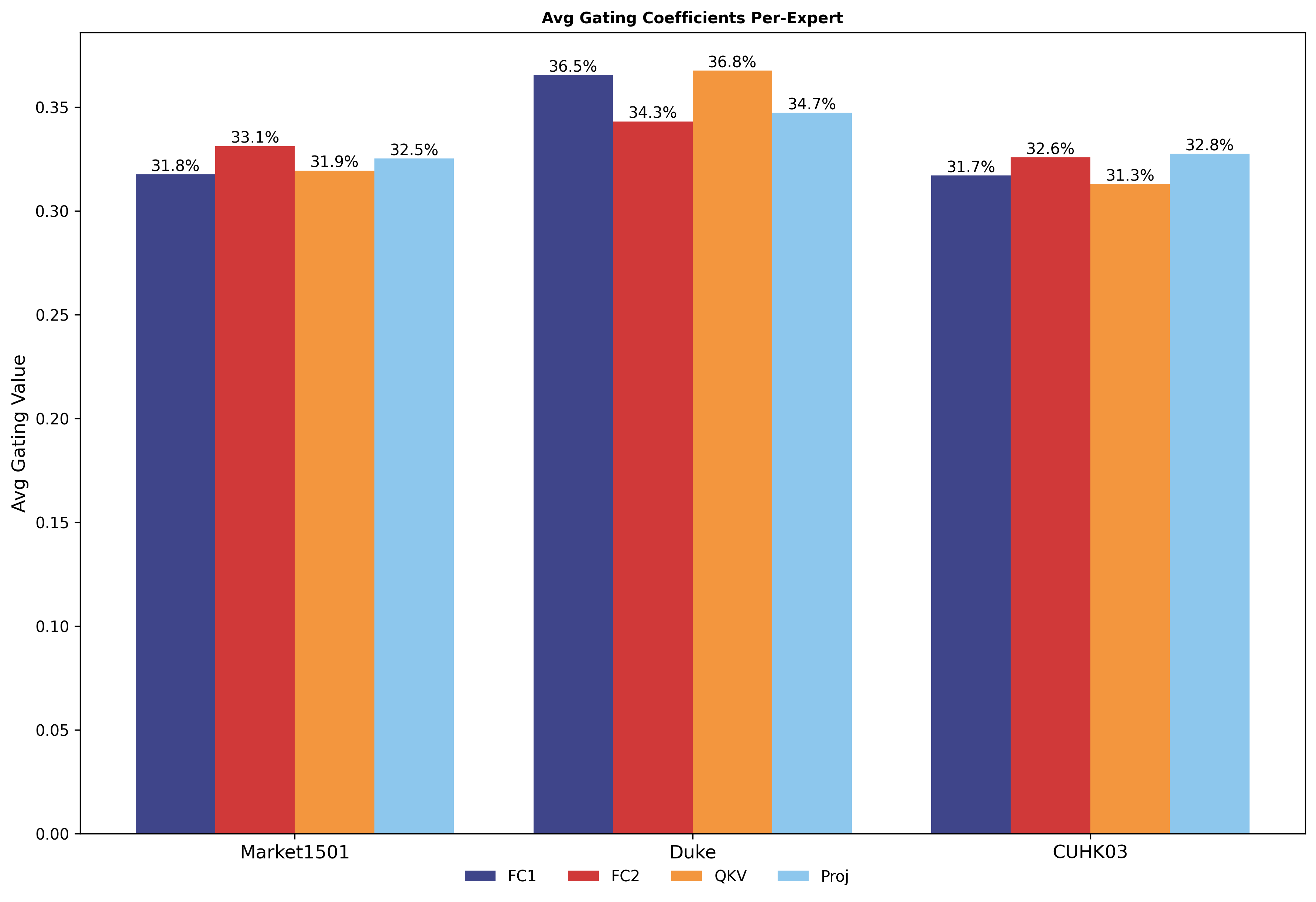}
    \subcaption{Average Average gating coefficient (Target: MSMT17)}
    \label{fig:weights:msmt}
  \end{subfigure}

  \vspace{2pt}
  \caption{Average gating coefficient $\bar g_{i,m}$ per target domain.}
  \label{fig:weights_w}
  \vspace{-6pt}
\end{figure*}

For a target domain instance \(x_t\), the gate outputs a weight $\alpha_{i,\ell,m}(x_t) \!\in\![0,1]$ for each source domain expert $i$ and layer $l$. Each Transformer layer contains four modules \(m\in\{\mathrm{FC1},\mathrm{FC2},\mathrm{QKV},\mathrm{Proj}\}\). Figure \ref{fig:weights_w} reports a single average per source expert and module by averaging across target instances and layers:
\[
\bar g_{i,m}
= \frac{1}{n_T\, L}\sum_{\ell=1}^{L}\;\sum_{t=1}^{n_T} \alpha_{i,\ell,m}(x_t),
\]
where \(n_T\) is the number of target instances in query set and \(L\) is the number of layers that contain module \(m\). 

\textbf{Findings from Fig.~\ref{fig:weights_w}} (i) \emph{Early modules are selective.} At QKV/FC1, distributions are peaked: for Duke and Market1501 targets, the MSMT17 expert typically receives the largest share, indicating that attention mixing and MLP expansion benefit most from the most diverse source. (ii) \textit{Late modules blend.} At Proj/FC2, weights are more uniform, showing consolidation of cues from multiple sources. (iii) \emph{Large‑target behavior.} On MSMT17, coefficients are generally even across experts, with only mild preference at QKV/FC1 and near‑uniform mixing at Proj/FC2. Overall, the gate allocates discriminative capacity in early modules and performs stabilizing mixing at later ones, consistent with composing per‑module residuals while keeping the backbone and adapters frozen.


\section{Analysis of Inter-Domain Synergies} 
Following the same notation of the previous section, instead of averaging, we concatenate all per-layer, per-instance gating weights and compute correlations on the pooled samples. For each module \(m\), we define $G_m \in \mathbb{R}^{n_TL \times S}$ as:
\[
[G_m]_{p,i} \;=\; \alpha_{i,\ell,m}(x^t).
\]
We then form the \(s\times s\) inter-source correlation matrix $C_m$ (i.e., the Pearson correlation across the \(n_TL\) concatenated samples), such that: 
\[
C_m \;=\; \mathrm{corr}(G_m).
\]

\begin{figure*}[t]
  \centering
  \begin{subfigure}{\textwidth}
    \centering
    \includegraphics[width=\textwidth]{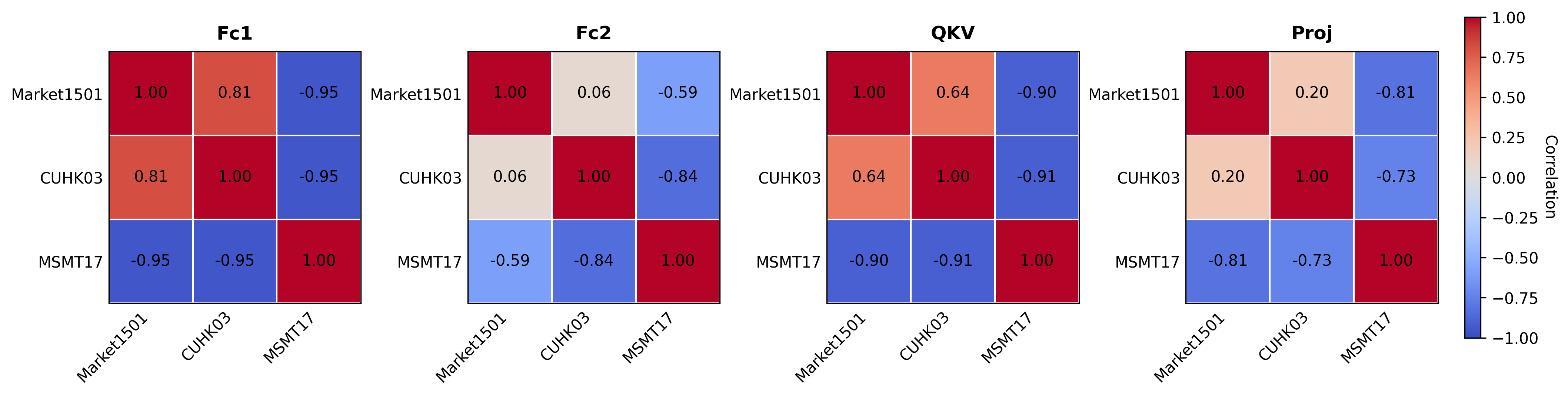}
    \subcaption{Target : DukeMTMC-reID}
    \label{fig:gate_corrs:a}
  \end{subfigure}
  \vspace{6pt}

  \begin{subfigure}{\textwidth}
    \centering
    \includegraphics[width=\textwidth]{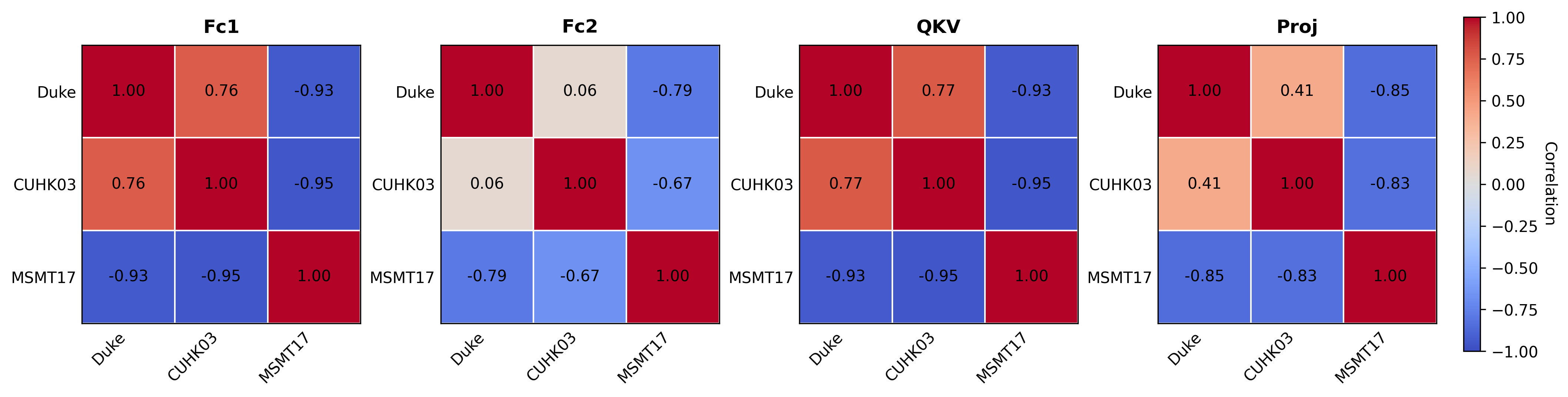}
    \subcaption{Target : Market1501}
    \label{fig:gate_corrs:b}
  \end{subfigure}
  \vspace{6pt}

  \begin{subfigure}{\textwidth}
    \centering
    \includegraphics[width=\textwidth]{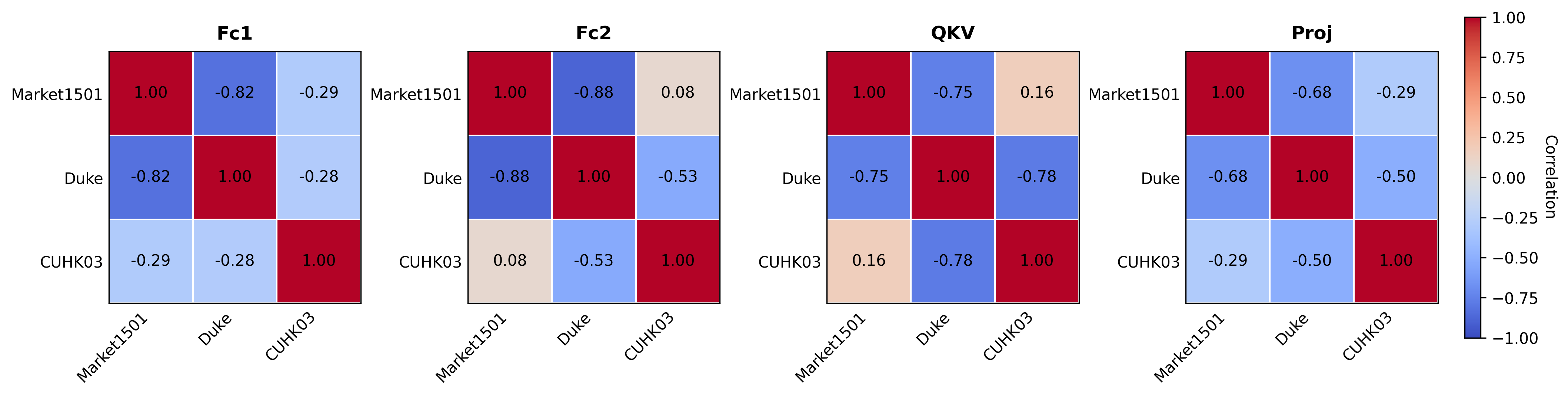}
    \subcaption{Target : MSMT17}
    \label{fig:gate_corrs:c}
  \end{subfigure}

  \vspace{-2pt}
  \caption{Overall gating-correlation visualization across layers.}
  \label{fig:gate_corr}
  \vspace{-6pt}
\end{figure*}

\textbf{Findings (Fig.~\ref{fig:gate_corr}).}
(i) \emph{Early modules are structured.} At QKV/FC1, correlations form clear “source teams”: for Duke as target, \{Market, CUHK03\} align and contrast with MSMT17; for Market1501, \{Duke, CUHK03\} align against MSMT17. For MSMT17, the dominant contrast is Market vs.\ Duke with CUHK03 weakly mediating. (ii) \emph{Late modules blend.} At Proj/FC2, correlations shrink toward $0$, indicating more uniform mixing that consolidates cues from multiple sources. Overall, the gate forms target‑dependent coalitions in early modules (routing discriminative capacity) and then stabilizes by blending in later modules, evidencing complementarity rather than redundancy among sources.


\section{Effect of Occlusion and Source Exclusion} Table \ref{tab:llo} evaluates two factors: (i) replacing Duke-DukeMTMC-reID (D) with the more challenging Occluded-DukeMTMC-reID \cite{Miao_2019_ICCV} (OC-D), and (ii) a leave-one-out source-exclusion study. We also include IUSTPersonReID (IUST)~\cite{moghaddam2025culturally}, which introduces a substantial domain shift. First, swapping D for OC-D consistently lowers performance (mAP/R-1), reflecting stronger occlusion and a larger mismatch to the target. Second, the LOO ablation shows that removing any source degrades results, confirming that all sources are complementary. Notably, although OC-D is individually weaker than the other source datasets, excluding it still causes a measurable drop, indicating that it contributes non-redundant information to the merge. In short, OC-D remains a useful component of the multi-source pool.


\begin{table}[ht]
\centering
\footnotesize 
\setlength{\tabcolsep}{3pt}        
\renewcommand{\arraystretch}{1} 
\begin{tabularx}{0.5\textwidth}{l|*{6}{>{\centering\arraybackslash}X}}
\hline
\multirow{2}{*}{\textbf{Methods}} &
\multicolumn{2}{c}{\textbf{D}} &
\multicolumn{2}{c}{\textbf{IUST}} &
\multicolumn{2}{c}{\textbf{M}} \\
\cline{2-7}
& \textbf{mAP} & \textbf{R1} & \textbf{mAP} & \textbf{R1} & \textbf{mAP} & \textbf{R1} \\
\hline
$\mathcal{L}_{\text{UDA}}$ + ViT\textsubscript{B/16} (LoRA) & 41.0 & 65.8 & 40.1 & 65.0 & 39.0 & 63.5 \\
\hline
\multicolumn{3}{l}{\textbf{\textit{Multi-Source (Duke + IUST + Market1501):}}} \\ \hline
SAGE-ReID (LOO = D)                 & \multicolumn{6}{c}{41.7 \quad 68.1} \\
SAGE-ReID (LOO = IUST)                 & \multicolumn{6}{c}{41.6 \quad 67.3} \\
SAGE-ReID (LOO = M)                 & \multicolumn{6}{c}{43.5 \quad 69.5} \\
Ties                       & \multicolumn{6}{c}{34.8 \quad 61.8} \\
\rowcolor{gray!15}
SAGE-ReID                  & \multicolumn{6}{c}{\textbf{44.5 \quad 70.2}} \\
\hline
\multirow{2}{*}{\textbf{Methods}} &
\multicolumn{2}{c}{\textbf{OC-D}} &
\multicolumn{2}{c}{\textbf{IUST}} &
\multicolumn{2}{c}{\textbf{M}} \\
\cline{2-7}
& \textbf{mAP} & \textbf{R1} & \textbf{mAP} & \textbf{R1} & \textbf{mAP} & \textbf{R1} \\
\hline
$\mathcal{L}_{\text{UDA}}$ + ViT\textsubscript{B/16} (LoRA) (Source Free)        & 34.2 & 61.7 & 40.1 & 65.0 & 39.0 & 63.5 \\
\hline
\multicolumn{3}{l}{\textbf{\textit{Multi-Source (OC-Duke + IUST + Market1501):}}} \\ \hline
SAGE-ReID (LOO = OC-D)                 & \multicolumn{6}{c}{41.7 \quad 68.1} \\
SAGE-ReID (LOO = IUST)                    & \multicolumn{6}{c}{39.5 \quad 65.6} \\
SAGE-ReID (LOO = M)              & \multicolumn{6}{c}{37.7 \quad 65.3} \\
\rowcolor{gray!15}
SAGE-ReID                 & \multicolumn{6}{c}{\textbf{42.1 \quad 68.3}} \\
\hline
\end{tabularx}
\caption{Impact of removing one source dataset contribution and adding Ocluded-DukeMTMC (OC-D). LOO stands for leave-one-out}
\label{tab:llo}
\vspace{-10pt}
\end{table}

\section{Feature Alignment Analysis}
Centered Kernel Alignment (CKA) \cite{kornblith2019similarity} measures representational similarity (1.0 = identical) between two models’ activations on the same inputs. We apply the same averaging procedure as in Sec. \ref{sec:E}, now on feature maps rather than gating weights. In Table \ref{tab:cka} we observe consistently high CKA between each single expert (M, CU, MS) and the gated mixture (M + CU + MS) on QKV/FC1/Proj (typically $0.90$–$0.96$), with lower scores at FC2 ($0.63$–$0.77$). This pattern indicates that the gating function largely preserves each expert’s representation in the QKV/FC1/Proj layers, while FC2 performs task-specific reshaping. The per-row means ($0.84$, $0.87$, $0.89$) further show that alignment holds across all experts, supporting that the gate mixes via soft selection rather than creating a new, conflicting representation.

\begin{table}[t]
\centering
\footnotesize
\setlength{\tabcolsep}{4pt}
\renewcommand{\arraystretch}{1.05}
\vspace{2pt}
\begin{tabularx}{\linewidth}{@{}l l >{\centering\arraybackslash}X >{\centering\arraybackslash}X >{\centering\arraybackslash}X >{\centering\arraybackslash}X >{\centering\arraybackslash}X@{}}
\toprule
\textbf{Target} & \textbf{Single / Combo} & \textbf{QKV} & \textbf{Proj} & \textbf{FC1} & \textbf{FC2} & \textbf{Mean} \\
\midrule
\multirow{3}{*}{\textbf{D}}
& M $\leftrightarrow$ (M+CU+MS)                 & 0.96 & 0.91 & 0.91 & 0.71 & 0.84 \\
& CU $\leftrightarrow$ (M+CU+MS)                 & 0.92 & 0.90 & 0.92 & 0.63 & 0.87 \\
& MS $\leftrightarrow$ (M+CU+MS)              & 0.95 & 0.92 & 0.93 & 0.77 & 0.89 \\
\bottomrule
\end{tabularx}
\caption{\textbf{Layer-wise CKA across experts/combinations.}
For \emph{Target Duke (D)}, linear CKA is computed by first averaging feature maps over test samples, then averaging across the 12 Transformer blocks for each layer (QKV, Proj, FC1, FC2).
\textbf{Mean} is the average across the four layers.
}
\label{tab:cka}
\vspace{-6pt}
\end{table}


{
    \small
    \bibliographystyle{ieeenat_fullname}
    \bibliography{references}

@String(CVPR= {IEEE Conf. Comput. Vis. Pattern Recog.})

@String(ICCV= {Int. Conf. Comput. Vis.})

@String(ECCV= {Eur. Conf. Comput. Vis.})

@String(TIP  = {IEEE Trans. Image Process.})

@String(ICLR = {Int. Conf. Learn. Represent.})

@String(PR   = {Pattern Recognition})

@String(AAAI = {AAAI})

@String(CVPR  = {CVPR})

@String(ICCV  = {ICCV})

@String(ECCV  = {ECCV})

@String(TIP   = {IEEE TIP})

@String(ICLR  = {ICLR})

@String(PR = {PR})

@InProceedings{Yang_2025_CVPR,
    author    = {Yang, Jinxi and Li, He and Du, Bo and Ye, Mang},
    title     = {Cheb-GR: Rethinking K-nearest Neighbor Search in Re-ranking for Person Re-identification},
    booktitle = {Proceedings of the Computer Vision and Pattern Recognition Conference (CVPR)},
    year      = {2025},
    pages     = {19261-19270}
}

@InProceedings{Zhang_2023_CVPR,
    author    = {Zhang, Guiwei and Zhang, Yongfei and Zhang, Tianyu and Li, Bo and Pu, Shiliang},
    title     = {PHA: Patch-Wise High-Frequency Augmentation for Transformer-Based Person Re-Identification},
    booktitle = {Proceedings of the IEEE/CVF Conference on Computer Vision and Pattern Recognition (CVPR)},
    year      = {2023},
    pages     = {14133-14142}
}

@InProceedings{He_2021_ICCV,
    author    = {He, Shuting and Luo, Hao and Wang, Pichao and Wang, Fan and Li, Hao and Jiang, Wei},
    title     = {TransReID: Transformer-Based Object Re-Identification},
    booktitle = {Proceedings of the IEEE/CVF International Conference on Computer Vision (ICCV)},
    year      = {2021},
    pages     = {15013-15022}
}

@InProceedings{Li_2021_CVPR,
    author    = {Li, Hanjun and Wu, Gaojie and Zheng, Wei-Shi},
    title     = {Combined Depth Space Based Architecture Search for Person Re-Identification},
    booktitle = {Proceedings of the IEEE/CVF Conference on Computer Vision and Pattern Recognition (CVPR)},
    year      = {2021},
    pages     = {6729-6738}
}

@InProceedings{Wei_2018_CVPR,
author = {Wei, Longhui and Zhang, Shiliang and Gao, Wen and Tian, Qi},
title = {Person Transfer GAN to Bridge Domain Gap for Person Re-Identification},
booktitle = {Proceedings of the IEEE Conference on Computer Vision and Pattern Recognition (CVPR)},
year = {2018}
}

@inproceedings{zheng2021exploiting,
  title={Exploiting sample uncertainty for domain adaptive person re-identification},
  author={Zheng, Kecheng and Lan, Cuiling and Zeng, Wenjun and Zhang, Zhizheng and Zha, Zheng-Jun},
  booktitle={Proceedings of the AAAI conference on Artificial Intelligence (AAAI)},
  volume={35},
  number={4},
  pages={3538--3546},
  year={2021}
}

@InProceedings{Zheng_2021_CVPR,
    author    = {Zheng, Kecheng and Liu, Wu and He, Lingxiao and Mei, Tao and Luo, Jiebo and Zha, Zheng-Jun},
    title     = {Group-aware Label Transfer for Domain Adaptive Person Re-identification},
    booktitle = {Proceedings of the IEEE/CVF Conference on Computer Vision and Pattern Recognition (CVPR)},
    year      = {2021},
    pages     = {5310-5319}
}

@InProceedings{Lee_2023_ICCV,
    author    = {Lee, Geon and Lee, Sanghoon and Kim, Dohyung and Shin, Younghoon and Yoon, Yongsang and Ham, Bumsub},
    title     = {Camera-Driven Representation Learning for Unsupervised Domain Adaptive Person Re-identification},
    booktitle = {Proceedings of the IEEE/CVF International Conference on Computer Vision (ICCV)},
    year      = {2023},
    pages     = {11453-11462}
}

@InProceedings{Dai_2021_ICCV,
    author    = {Dai, Yongxing and Liu, Jun and Sun, Yifan and Tong, Zekun and Zhang, Chi and Duan, Ling-Yu},
    title     = {IDM: An Intermediate Domain Module for Domain Adaptive Person Re-ID},
    booktitle = {Proceedings of the IEEE/CVF International Conference on Computer Vision (ICCV)},
    year      = {2021},
    pages     = {11864-11874}
}

@inproceedings{ge2020mutual,
  title={Mutual Mean-Teaching: Pseudo Label Refinery for Unsupervised Domain Adaptation on Person Re-identification},
  author={Ge, Yixiao and Chen, Dapeng and Li, Hongsheng},
  booktitle={International Conference on Learning Representations (ICLR)},
  year={2020},
}

@inproceedings{ge2020selfpaced,
    title={Self-paced Contrastive Learning with Hybrid Memory for Domain Adaptive Object Re-ID},
    author = {Ge, Yixiao and Zhu, Feng and Chen, Dapeng and Zhao, Rui and Li, hongsheng},
    booktitle={Advances in Neural Information Processing Systems (NeurIPS)},
    year={2020}
}

@inproceedings{he2022secret,
  title={Secret: Self-consistent pseudo label refinement for unsupervised domain adaptive person re-identification},
  author={He, Tao and Shen, Leqi and Guo, Yuchen and Ding, Guiguang and Guo, Zhenhua},
  booktitle={Proceedings of the AAAI conference on Artificial Intelligence (AAAI)},
  volume={36},
  number={1},
  pages={879--887},
  year={2022}
}

@InProceedings{Isobe_2021_ICCV,
    author    = {Isobe, Takashi and Li, Dong and Tian, Lu and Chen, Weihua and Shan, Yi and Wang, Shengjin},
    title     = {Towards Discriminative Representation Learning for Unsupervised Person Re-Identification},
    booktitle = {Proceedings of the IEEE/CVF International Conference on Computer Vision (ICCV)},
    year      = {2021},
    pages     = {8526-8536}
}

@inproceedings{li2022reliability,
  title={Reliability exploration with self-ensemble learning for domain adaptive person re-identification},
  author={Li, Zongyi and Shi, Yuxuan and Ling, Hefei and Chen, Jiazhong and Wang, Qian and Zhou, Fengfan},
  booktitle={Proceedings of the AAAI conference on Artificial Intelligence (AAAI)}, 
  volume={36},
  number={2},
  pages={1527--1535},
  year={2022}
}

@InProceedings{zhao2020unsupervised,
    author    = {Zhao, Fang and Liao, Shengcai and Xie, Guo-Sen and Zhao, Jian and Zhang, Kaihao and Shao, Ling},
    title     = {Unsupervised Domain Adaptation with Noise Resistible Mutual-Training for Person Re-Identification},
    booktitle = {Proceedings of the European Conference on Computer Vision (ECCV)},
    year      = {2020},
    pages     = {526-544}
}

@inproceedings{zheng2021online,
  title={Online pseudo label generation by hierarchical cluster dynamics for adaptive person re-identification},
  author={Zheng, Yi and Tang, Shixiang and Teng, Guolong and Ge, Yixiao and Liu, Kaijian and Qin, Jing and Qi, Donglian and Chen, Dapeng},
  booktitle = {Proceedings of the IEEE/CVF International Conference on Computer Vision (ICCV)},
  pages={8371--8381},
  year={2021}
}

@InProceedings{hu2022lora,
    author    = {Hu, Edward J. and Shen, Yelong and Wallis, Phillip and Allen-Zhu, Zeyuan and Li, Yuanzhi and Wang, Shean and Wang, Lu and Chen, Weizhu},
    title     = {LoRA: Low-Rank Adaptation of Large Language Models},
    booktitle = {Proceedings of the International Conference on Learning Representations (ICLR)},
    year      = {2022}
}

@article{fan2018unsupervised,
  title={Unsupervised person re-identification: Clustering and fine-tuning},
  author={Fan, Hehe and Zheng, Liang and Yan, Chenggang and Yang, Yi},
  journal={ACM Transactions on Multimedia Computing, Communications, and Applications (TOMM)},
  volume={14},
  number={4},
  pages={1--18},
  year={2018},
  publisher={ACM New York, NY, USA}
}

@InProceedings{Zheng_2015_ICCV,
author = {Zheng, Liang and Shen, Liyue and Tian, Lu and Wang, Shengjin and Wang, Jingdong and Tian, Qi},
title = {Scalable Person Re-Identification: A Benchmark},
booktitle = {Proceedings of the IEEE/CVF International Conference on Computer Vision (ICCV)},
month = {December},
year = {2015}
}

@InProceedings{ristani2016performance,
    author    = {Ristani, Ergys and Solera, Francesco and Zou, Roger and Cucchiara, Rita and Tomasi, Carlo},
    title     = {Performance Measures and a Data Set for Multi-Target, Multi-Camera Tracking},
    booktitle = {Proceedings of the European Conference on Computer Vision (ECCV) Workshops},
    year      = {2016},
    pages     = {17-35}
}

@InProceedings{Li_2014_CVPR,
author = {Li, Wei and Zhao, Rui and Xiao, Tong and Wang, Xiaogang},
title = {DeepReID: Deep Filter Pairing Neural Network for Person Re-Identification},
booktitle = {Proceedings of the IEEE Conference on Computer Vision and Pattern Recognition (CVPR)},
month = {June},
year = {2014}
}

@inproceedings{dosovitskiy2020image,
title={An Image is Worth 16x16 Words: Transformers for Image Recognition at Scale},
author={Alexey Dosovitskiy and Lucas Beyer and Alexander Kolesnikov and Dirk Weissenborn and Xiaohua Zhai and Thomas Unterthiner and Mostafa Dehghani and Matthias Minderer and Georg Heigold and Sylvain Gelly and Jakob Uszkoreit and Neil Houlsby},
booktitle={International Conference on Learning Representations (ICLR)},
year={2021}
}

@InProceedings{zhong2020random,
    author    = {Zhong, Zhun and Zheng, Liang and Kang, Guoliang and Li, Shaozi and Yang, Yi},
    title     = {Random Erasing Data Augmentation},
    booktitle = {Proceedings of the AAAI Conference on Artificial Intelligence (AAAI)},
    year      = {2020},
    volume    = {34},
    number    = {7},
    pages     = {13001-13008}
}

@article{xian2025distilling,
  title={Distilling consistent relations for multi-source domain adaptive person re-identification},
  author={Xian, Yuqiao and Peng, Yi-Xing and Sun, Xing and Zheng, Wei-Shi},
  journal={Pattern Recognition (PR)},
  volume={157},
  pages={110821},
  year={2025}
}

@InProceedings{Peng_2019_ICCV,
author = {Peng, Xingchao and Bai, Qinxun and Xia, Xide and Huang, Zijun and Saenko, Kate and Wang, Bo},
title = {Moment Matching for Multi-Source Domain Adaptation},
booktitle = {Proceedings of the IEEE/CVF International Conference on Computer Vision (ICCV)},
month = {October},
year = {2019}
}

@InProceedings{ganin2015unsupervised,
    author    = {Ganin, Yaroslav and Lempitsky, Victor},
    title     = {Unsupervised Domain Adaptation by Backpropagation},
    booktitle = {Proceedings of the International Conference on Machine Learning (ICML)},
    year      = {2015},
    pages     = {1180-1189}
}

@InProceedings{yang2020curriculum,
    author    = {Yang, Luyu and Balaji, Yogesh and Lim, Ser-Nam and Shrivastava, Abhinav},
    title     = {Curriculum Manager for Source Selection in Multi-Source Domain Adaptation},
    booktitle = {Proceedings of the European Conference on Computer Vision (ECCV)},
    year      = {2020},
    pages     = {608-624}
}

@InProceedings{Chang_2019_CVPR,
    author    = {Chang, Woong-Gi and You, Tackgeun and Seo, Seonguk and Kwak, Suha and Han, Bohyung},
    title     = {Domain-Specific Batch Normalization for Unsupervised Domain Adaptation},
    booktitle = {Proceedings of the IEEE/CVF Conference on Computer Vision and Pattern Recognition (CVPR)},
    year      = {2019},
    pages     = {7354-7362}
}

@inproceedings{wu2024mixture,
  title={Mixture of LoRA experts},
  author={Wu, Xun and Huang, Shaohan and Wei, Furu},
  booktitle={International Conference on Learning Representations (ICLR)},
  year={2024}
}

@InProceedings{Zhu_2022_CVPR,
    author    = {Zhu, Haowei and Ke, Wenjing and Li, Dong and Liu, Ji and Tian, Lu and Shan, Yi},
    title     = {Dual Cross-Attention Learning for Fine-Grained Visual Categorization and Object Re-Identification},
    booktitle = {Proceedings of the IEEE/CVF Conference on Computer Vision and Pattern Recognition (CVPR)},
    month     = {June},
    year      = {2022},
    pages     = {4692-4702}
}

@article{zhang2021seeing,
  title={Seeing like a human: Asynchronous learning with dynamic progressive refinement for person re-identification},
  author={Zhang, Quan and Lai, Jianhuang and Feng, Zhanxiang and Xie, Xiaohua},
  journal={IEEE Transactions on Image Processing (TIP)},
  volume={31},
  pages={352--365},
  year={2021}
}

@InProceedings{Bai_2021_CVPR,
    author    = {Bai, Zechen and Wang, Zhigang and Wang, Jian and Hu, Di and Ding, Errui},
    title     = {Unsupervised Multi-Source Domain Adaptation for Person Re-Identification},
    booktitle = {Proceedings of the IEEE/CVF Conference on Computer Vision and Pattern Recognition (CVPR)},
    year      = {2021},
    pages     = {12914-12923}
}

@InProceedings{Deng_2018_CVPR,
author = {Deng, Weijian and Zheng, Liang and Ye, Qixiang and Kang, Guoliang and Yang, Yi and Jiao, Jianbin},
title = {Image-Image Domain Adaptation With Preserved Self-Similarity and Domain-Dissimilarity for Person Re-Identification},
booktitle = {Proceedings of the IEEE Conference on Computer Vision and Pattern Recognition (CVPR)},
month = {June},
year = {2018}
}

@InProceedings{Goodfellow_2014_NeurIPS,
    author    = {Goodfellow, Ian J. and Pouget-Abadie, Jean and Mirza, Mehdi and Xu, Bing and Warde-Farley, David and Ozair, Sherjil and Courville, Aaron and Bengio, Yoshua},
    title     = {Generative Adversarial Nets},
    booktitle = {Advances in Neural Information Processing Systems (NeurIPS)},
    year      = {2014},
    pages     = {2672--2680}
}

@article{kodali2017convergence,
  title={On convergence and stability of gans},
  author={Kodali, Naveen and Abernethy, Jacob and Hays, James and Kira, Zsolt},
  journal={arXiv preprint arXiv:1705.07215},
  year={2017}
}

@InProceedings{Zheng_2021_ICCV,
    author    = {Zheng, Yi and Tang, Shixiang and Teng, Guolong and Ge, Yixiao and Liu, Kaijian and Qin, Jing and Qi, Donglian and Chen, Dapeng},
    title     = {Online Pseudo Label Generation by Hierarchical Cluster Dynamics for Adaptive Person Re-Identification},
    booktitle = {Proceedings of the IEEE/CVF International Conference on Computer Vision (ICCV)},
    year      = {2021},
    pages     = {8371-8381}
}

@InProceedings{verma2019manifold,
    author    = {Verma, Vikas and Lamb, Alex and Beckham, Christopher and Najafi, Amir and Mitliagkas, Ioannis and Lopez-Paz, David and Bengio, Yoshua},
    title     = {Manifold Mixup: Better Representations by Interpolating Hidden States},
    booktitle = {Proceedings of the International Conference on Machine Learning (ICML)},
    year      = {2019},
    pages     = {6438-6447}
}

@Article{buehler2024x,
    author    = {Buehler, Eric L. and Buehler, Markus J.},
    title     = {X-LoRA: Mixture of Low-Rank Adapter Experts, a Flexible Framework for Large Language Models with Applications in Protein Mechanics and Molecular Design},
    journal   = {APL Machine Learning},
    year      = {2024},
    volume    = {2},
    number    = {2},
    pages     = {026119}
}

@InProceedings{feng2024mixture,
    author    = {Feng, Wenfeng and Hao, Chuzhan and Zhang, Yuewei and Han, Yu and Wang, Hao},
    title     = {Mixture-of-LoRAs: An Efficient Multitask Tuning Method for Large Language Models},
    booktitle = {Proceedings of the Joint International Conference on Computational Linguistics, Language Resources and Evaluation (LREC-COLING)},
    year      = {2024},
    pages     = {11371-11380}
}

@article{luo2024moelora,
  title={Moelora: Contrastive learning guided mixture of experts on parameter-efficient fine-tuning for large language models},
  author={Luo, Tongxu and Lei, Jiahe and Lei, Fangyu and Liu, Weihao and He, Shizhu and Zhao, Jun and Liu, Kang},
  journal={arXiv preprint arXiv:2402.12851},
  year={2024}
}

@InProceedings{muqeeth2024learning,
    author    = {Muqeeth, Mohammed and Liu, Haokun and Liu, Yufan and Raffel, Colin},
    title     = {Learning to Route Among Specialized Experts for Zero-Shot Generalization},
    booktitle = {Proceedings of the International Conference on Machine Learning (ICML)},
    year      = {2024},
    pages     = {1496-1513}
}

@InProceedings{yu2024language,
    author    = {Yu, Le and Yu, Bowen and Yu, Haiyang and Huang, Fei and Li, Yongbin},
    title     = {Language Models Are Super Mario: Absorbing Abilities from Homologous Models as a Free Lunch},
    booktitle = {Proceedings of the International Conference on Machine Learning (ICML)},
    year      = {2024}
}

@inproceedings{huang2023lorahub,
title={LoraHub: Efficient Cross-Task Generalization via Dynamic Lo{RA} Composition},
author={Chengsong Huang and Qian Liu and Bill Yuchen Lin and Tianyu Pang and Chao Du and Min Lin},
booktitle={Conference on Language Modeling (COLM)},
year={2024}
}

@article{ye2021deep,
  title={Deep learning for person re-identification: A survey and outlook},
  author={Ye, Mang and Shen, Jianbing and Lin, Gaojie and Xiang, Tao and Shao, Ling and Hoi, Steven CH},
  journal={IEEE Transactions on Pattern Analysis and Machine Intelligence (TPAMI)},
  volume={44},
  number={6},
  pages={2872--2893},
  year={2021}
}

@article{khan2024deep,
  title={Deep-ReID: Deep features and autoencoder assisted image patching strategy for person re-identification in smart cities surveillance},
  author={Khan, Samee Ullah and Hussain, Tanveer and Ullah, Amin and Baik, Sung Wook},
  journal={Multimedia Tools and Applications},
  volume={83},
  number={5},
  pages={15079--15100},
  year={2024}
}

@InProceedings{wong2020identifying,
    author    = {Wong, Kelvin and Wang, Shenlong and Ren, Mengye and Liang, Ming and Urtasun, Raquel},
    title     = {Identifying Unknown Instances for Autonomous Driving},
    booktitle = {Proceedings of the Conference on Robot Learning (CoRL)},
    year      = {2020},
    pages     = {384-393}
}

@article{yadav2023ties,
  title={Ties-merging: Resolving interference when merging models},
  author={Yadav, Prateek and Tam, Derek and Choshen, Leshem and Raffel, Colin A and Bansal, Mohit},
  journal={Advances in Neural Information Processing Systems},
  volume={36},
  pages={7093--7115},
  year={2023}
}

@inproceedings{zheng2021group,
  title={Group-aware label transfer for domain adaptive person re-identification},
  author={Zheng, Kecheng and Liu, Wu and He, Lingxiao and Mei, Tao and Luo, Jiebo and Zha, Zheng-Jun},
  booktitle={Proceedings of the IEEE/CVF conference on computer vision and pattern recognition},
  pages={5310--5319},
  year={2021}
}

@InProceedings{Guichemerre_2024_CVPR,
    author    = {Guichemerre, Alexis and Belharbi, Soufiane and Mayet, Tsiry and Murtaza, Shakeeb and Shamsolmoali, Pourya and Mccaffrey, Luke and Granger, Eric},
    title     = {Source-Free Domain Adaptation of Weakly-Supervised Object Localization Models for Histology},
    booktitle = {Proceedings of the IEEE/CVF Conference on Computer Vision and Pattern Recognition (CVPR) Workshops},
    month     = {June},
    year      = {2024},
    pages     = {33-43}
}

@article{ilharco2022editing,
  title={Editing models with task arithmetic},
  author={Ilharco, Gabriel and Ribeiro, Marco Tulio and Wortsman, Mitchell and Gururangan, Suchin and Schmidt, Ludwig and Hajishirzi, Hannaneh and Farhadi, Ali},
  journal={arXiv preprint arXiv:2212.04089},
  year={2022}
}

@inproceedings{NEURIPS2023_1644c9af,
 author = {Yadav, Prateek and Tam, Derek and Choshen, Leshem and Raffel, Colin A and Bansal, Mohit},
 booktitle = {Advances in Neural Information Processing Systems},
 editor = {A. Oh and T. Naumann and A. Globerson and K. Saenko and M. Hardt and S. Levine},
 pages = {7093--7115},
 publisher = {Curran Associates, Inc.},
 title = {TIES-Merging: Resolving Interference When Merging Models},
 url = {https://proceedings.neurips.cc/paper_files/paper/2023/file/1644c9af28ab7916874f6fd6228a9bcf-Paper-Conference.pdf},
 volume = {36},
 year = {2023}
}

@article{stoica2024model,
  title={Model merging with svd to tie the knots},
  author={Stoica, George and Ramesh, Pratik and Ecsedi, Boglarka and Choshen, Leshem and Hoffman, Judy},
  journal={arXiv preprint arXiv:2410.19735},
  year={2024}
}

@InProceedings{Gargiulo_2025_CVPR,
    author    = {Gargiulo, Antonio Andrea and Crisostomi, Donato and Bucarelli, Maria Sofia and Scardapane, Simone and Silvestri, Fabrizio and Rodol\`a, Emanuele},
    title     = {Task Singular Vectors: Reducing Task Interference in Model Merging},
    booktitle = {Proceedings of the IEEE/CVF Conference on Computer Vision and Pattern Recognition (CVPR)},
    month     = {June},
    year      = {2025},
    pages     = {18695-18705}
}

@inproceedings{kornblith2019similarity,
  title={Similarity of neural network representations revisited},
  author={Kornblith, Simon and Norouzi, Mohammad and Lee, Honglak and Hinton, Geoffrey},
  booktitle={International conference on machine learning},
  pages={3519--3529},
  year={2019},
  organization={PMlR}
}

@InProceedings{Miao_2019_ICCV,
author = {Miao, Jiaxu and Wu, Yu and Liu, Ping and Ding, Yuhang and Yang, Yi},
title = {Pose-Guided Feature Alignment for Occluded Person Re-Identification},
booktitle = {Proceedings of the IEEE/CVF International Conference on Computer Vision (ICCV)},
month = {October},
year = {2019}
}

@article{moghaddam2025culturally,
  title={A culturally-aware benchmark for Person Re-Identification in modest attire},
  author={Moghaddam, Alireza Sedighi and Anvari, Fatemeh and Haghighi, Mohammadjavad Mirshekari and Fakhari, Mohammadali and Mohammadi, Mohammad Reza},
  journal={Engineering Applications of Artificial Intelligence},
  volume={158},
  pages={111494},
  year={2025},
  publisher={Elsevier}
}
}

\end{document}